%% file: main.tex
\documentclass{article}

\usepackage[nonatbib,final]{neurips_2021}
\usepackage{soul}

\input{custom-notations}

\usepackage[utf8]{inputenc} % allow utf-8 input
\usepackage[T1]{fontenc}    % use 8-bit T1 fonts
\usepackage{hyperref}       % hyperlinks
\usepackage{url}            % simple URL typesetting
\usepackage{booktabs}       % professional-quality tables
\usepackage{amsfonts}       % blackboard math symbols
\usepackage{nicefrac}       % compact symbols for 1/2, etc.
\usepackage{microtype}      % microtypography
\usepackage{xcolor}         % colors
\usepackage{enumitem}  %to handle lists

\title{Model, sample, and epoch-wise descents: exact solution of gradient flow in the random feature model}

\author{%
  Antoine Bodin \qquad \& \qquad Nicolas Macris%\thanks{Use footnote for providing further information
  \\\\
  Communication Theory Laboratory,\\ 
  School of Computer and Communication Sciences,\\
  Ecole Polytechnique F\'ed\'erale de Lausanne\\
  \texttt{antoine.bodin@epfl.ch},\\
  \texttt{nicolas.macris@epfl.ch} \\
}

\begin{document}

\maketitle

\begin{abstract}
Recent evidence has shown the existence of a so-called double-descent and even triple-descent behavior for the generalization error of deep-learning models. 
This important phenomenon commonly appears in implemented neural network architectures, and also seems to emerge in epoch-wise curves during the training process.
A recent line of research has highlighted that random matrix tools can be used to obtain precise analytical asymptotics of the generalization (and training) errors of the random feature model. 
In this contribution, we analyze the {\it whole temporal behavior} of the generalization and training errors under gradient flow for the random feature model. We show that in the asymptotic limit of large system size the {\it full time-evolution} path of both errors can be calculated analytically. This allows us to observe how the double and triple descents develop over time, if and when early stopping is an option, and also observe time-wise descent structures. Our techniques are based on Cauchy complex integral representations of the errors together with recent random matrix methods based on linear pencils. 
\end{abstract}

\input{main-introduction}

%\section{related literatures} % à enlever
%cite{ADVANI2020428}

\input{main-theory}

\section{Results and insights}\label{results-and-insights}
\input{main-result}

\input{main-insights}

\input{main-sketch-proof}

\input{main-conclusion}

%\section*{Acknowledgment} 
\acksection
The work of A. B has been supported by Swiss National Science Foundation grant no 200020 182517.

\newpage 

\bibliographystyle{unsrt}
%\small
\bibliography{bibliography}

\newpage

\newpage
\appendix

\input{app-test-error}
\input{app-cauchy-integral-representation}
\input{app-random-matrix-limits}
\input{app-linear-pencil}
\input{app-replica}
\input{app-numerical-experiments}

\end{document}

%% file: custom-notations.tex
\usepackage{graphicx}
\usepackage{amsfonts}
\usepackage{mathtools,amssymb}
\usepackage{graphicx}
\usepackage{float}
\usepackage{amsmath}
\usepackage{subfig}
\usepackage{float}
\usepackage[super]{nth}

\usepackage{tikz}
\usetikzlibrary{positioning}
\tikzset{%
  every neuron/.style={
    circle,
    draw,
    minimum size=1cm
  },
  neuron missing/.style={
    draw=none, 
    scale=2,
    text height=0.333cm,
    execute at begin node=\color{black}$\vdots$
  },
}

\newcommand{\norm}[1]{\left\lVert#1\right\rVert}

\newtheorem{theorem}{Theorem}[section]
\newtheorem{proposition}{Proposition}[theorem]

\newtheorem{definition}{Definition}[section]
\newtheorem{assumptions}{Assumptions}[section]

\newtheorem{result}{Result}[section]

\newcommand{\opL}{\mathcal L}
\newcommand{\opR}{\mathcal R}
\renewcommand{\Re}{\operatorname{Re}}
\renewcommand{\Im}{\operatorname{Im}}
\newcommand{\mean}[1]{\mathbb E#1}
\newcommand{\cov}{\text{Cov}}
\newcommand{\trace}[1]{\text{Tr}\left[#1\right]}
\newcommand{\traceLim}[2][17.5]{ \text{Tr}_{#1}\left[#2\right] }
\newcommand{\dd}[1]{\mathrm{d}#1}
\newcommand{\Real}{\mathbb R}
\newcommand{\Complex}{\mathbb C}
\newcommand{\Sp}{\text{Sp} }

\newcommand{\Normal}{\mathcal N}

\newcommand{\MSEtrain}{\mathcal H^{\text{train}}}
\newcommand{\MSEtest}{ \mathcal H^{\text{test}}}

\newcommand{\barMSEtrain}{\bar{\mathcal H}^{\text{train}}}
\newcommand{\barMSEtest}{\bar{\mathcal H}^{\text{test}}}
\newcommand{\Zlin}{Z_{\text{lin}}}

%% file: main-introduction.tex
\section{Introduction}\label{sec:intro}
Deep learning models have vastly increased in terms of number of parameters in the architecture and data sample sizes with recent applications using unprecedented numbers with as much as 175 billions parameters trained over billions of tokens \cite{brown2020language}. Such massive amounts of data and growing training budgets have spurred research seeking empirical power laws to scale model sizes appropriately with available resources \cite{kaplan2020scaling}, and nowadays it is common wisdom among practitioners that "larger models are better". This ongoing trend has been challenging classical statistical modeling where it is thought that increasing the number of parameters past an interpolation threshold (at which the training error vanishes while the test error usually increases) is doomed to over-fit the data \cite{hastie01statisticallearning}. We refer to \cite{DBLP:journals/corr/ZhangBHRV16} for a recent extensive discussion on this contradictory state of affairs.
%, and it thus seems wise to keep the model size moderate with respect to data size 
%ANCIENNE PHRASE: Still, the on-going trend shows both data sizes and sample sizes of roughly the same order of magnitude, {\color{blue} with a staggering practice from a statistical modelling perspective}  where it is commonly thought that increasing number of parameters is often doomed to over-fitting the data, and it seems wise to keep it low with respect to the size of the data. 
%FROM KAPLAN ET AL: These relationships allow us to determine the optimal allocation of a fixed compute budget. Larger models are significantly more sample-efficient, such that optimally compute-efficient training involves training very large models on a relatively modest amount of data and stopping significantly before convergence
Progress towards rationalizing this situation came from a series of papers \cite{BelkinPNAS2019, pmlr-v80-belkin18a, pmlr-v89-belkin19a, Belkin_2020, Spigler_2019, Geiger_2020, ADVANI2020428} evidencing the existence of phases
where increasing the number of parameters beyond the interpolation threshold can actually achieve good generalization, and the characteristic ${\rm U}$ curve of the bias-variance tradeoff is followed by a "descent" of the generalization error.  This phenomenon has been called the {\it double descent} and was analytically corroborated in linear models \cite{Hastie-Montanari-2019, NEURIPS2020_37740d59, Sahai-et-al, Bartlett30063, Thrampoulidis2021} as well as random feature (RF) (or random feature regression) shallow network models \cite{mei2020generalization, liao:hal-02971807, pmlr-v119-gerace20a, pmlr-v119-d-ascoli20a}.
Many of these works provide rigorous proofs with precise asymptotic expressions of double descent curves. Further developments have brought forward rich phenomenology, for example, a triple-descent phenomenon \cite{dascoli2020triple} linked to the degree of non-linearity of the activation function.
Further empirical evidence \cite{nakkiran2019deep} has also shown that  a similar effect occurs {\it while} training (ResNet18s on CIFAR10 trained using Adam) and has been called {\it epoch-wise double descent}. Moreover the authors of \cite{nakkiran2019deep} extensively test various CIFAR data sets, architectures (CNNs, ResNets, Transformers) and optimizers (SGD, Adam) and 
classify their observations into three types of double descents: (i) model-wise double descent when the number of network parameters is varied; (ii) sample-wise double descent when the data set size is varied; and (iii) epoch-wise double descent which occurs while training.  We wish to note that sample-wise double descent was derived long ago in precursor work on single layer perceptron networks \cite{Opper1995,engelvandenbroeck2001}.
An important theoretical challenge is to unravel all these structures in a unified analytical way and understand how generalization error evolves in time.

In this contribution we achieve a detailed analytical analysis of the gradient flow dynamics of the RF model (or regression) in the high-dimensional 
asymptotic limit. The model was initially introduced in \cite{RechtRahimi2007} as an approximation of kernel machines; more recently it has been recognized as an important playground for theoretical analysis of the model-wise double descent phenomenon, using tools from random matrix theory \cite{mei2020generalization, liao:hal-02971807, pmlr-v119-jacot20a}. Following \cite{mei2020generalization} we view the RF model 
as a $2$-layer neural network with fixed-random-first-layer-weights and dynamical second layer learned weights. The data is given by $n$ training pairs constituted of $d$-dimensional input vectors and output given by a linear function with additive gaussian noise.  The data is fed through $N$ neurons with a non-linear activation function and followed by one linear neuron whose weights we learn by gradient descent over a quadratic loss function. The high-dimensional asymptotic limit is defined as the regime $n,d,N \to +\infty$ while the ratios tend to finite values $\frac{N}{d} \to \psi$ and $\frac{n}{d} \to \phi$. 
As the training loss is convex one expects that the least-squares predictor (with Moore-Penrose inversion) gives the long time behavior of gradient descent. This has led to the calculation of highly non-trivial analytical algebraic expressions for training and generalization errors which describe (model-wise and sample-wise) double and triple descent curves \cite{mei2020generalization, dascoli2020triple}. However, to the best of our knowledge, there is no complete analytical derivation of the whole time evolution of the two errors.

\begin{figure}[h]
  \centering
  \subfloat{
    \includegraphics[width=0.4\textwidth]{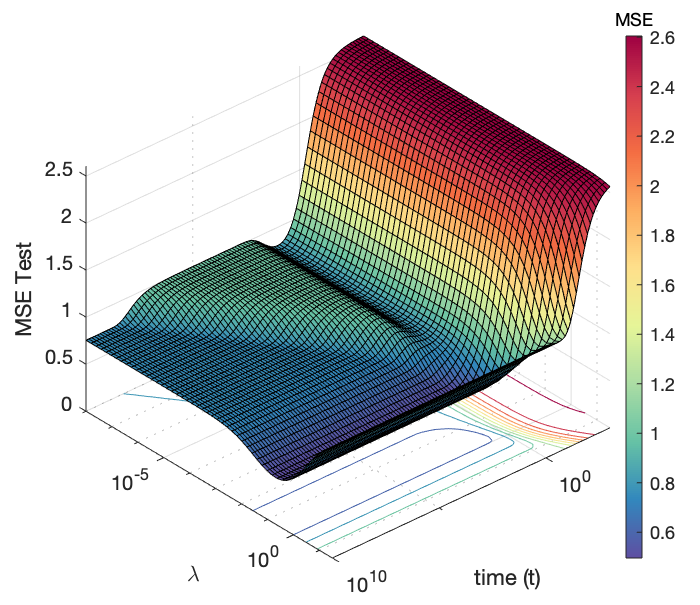}
  }\qquad
  \subfloat{
    \includegraphics[width=0.415\textwidth]{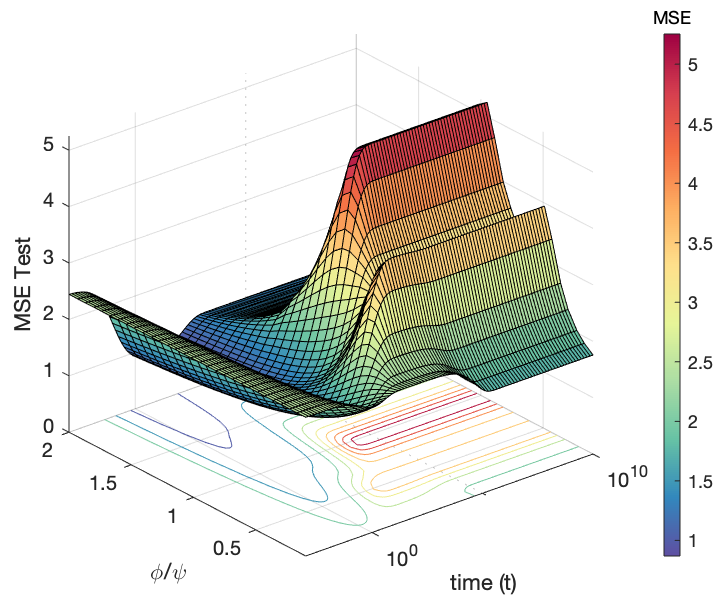}
  }
  \caption{\small 
  3D plot of analytical test error evolution. See Figs. \ref{fig:time_double_descent} and \ref{fig:time-tripledescent} (on the right) for parameter values.
  %Analytical training error and test error evolution with parameters
  %$(\mu,\nu,\psi,r,s,\lambda) = (0.9,0.1,2,1,0.8,0.0001)$. Parameters $\mu$, $\nu$ characterize the activation function $\sigma$. See Sect. \ref{subsec:modeldescription}.
  }
  \label{fig:intro}
\end{figure}

We analyze the gradient flow equations in the high-dimensional regime and deduce the whole time evolution of the training and generalization errors. Numerical simulations show that the gradient flow is an excellent approximation of gradient descent in the high-dimensional regime as long as the step size is small enough (see Fig. \ref{fig:tripledescent}). Main contributions presented in detail in Sect. \ref{results-and-insights} comprise:

{\bf a.} Results \ref{th:main} and \ref{result-infinite-time} give expressions of the time evolution of the errors in terms of {\it one and two-dimensional integrals over spectral densities whose Stieltjes transforms are given by a closed set of purely algebraic equations}. The expressions lend themselves to numerical computation as illustrated in Fig. \ref{fig:intro} and more extensively in Sect. \ref{results-and-insights} and the supplementary material.

{\bf b.} Model and sample-wise double descents develop after some definite time at the interpolation threshold and are preceded by a {\it dip or minimum} before the spike develops. This indicates that early stopping is beneficial for some parameter regimes. A similar behavior also occurs for the triple descent. (See Figs. \ref{fig:time_double_descent0}, \ref{fig:time-tripledescent} and the 3D version Fig. \ref{fig:intro}).

{\bf c.} We observe two kinds of epoch-wise "descent" structures. The first is a {\it double plateau} monotonously descending structure at widely different time scales in the largely overparameterized regime (see Fig. \ref{fig:time_double_descent0}). The second is an {\it epoch-wise double descent} similar to the one found in \cite{nakkiran2019deep}. In fact, as in \cite{nakkiran2019deep}, rather than a spike, this double descent appears to be an {\it elongated bump} over a wide time scale (see Fig. \ref{fig:time_double_descent} and the 3D version Fig. \ref{fig:intro}).

Let us say a few words about the techniques used in this work. We first translate the gradient flow equations for the learned weights of the second layer into a set of integro-differential equations for generating functions, as in \cite{pmlr-v134-bodin21a}, involving the resolvent of a random matrix (constructed out of the fixed first layer weights, the data, and the non-linear activation). The solution of the integro-differential equations and the time evolution of the errors can then be expressed in terms of Cauchy complex integral representation which has the advantage to decouple the time dependence and static contributions involving traces of algebraic combinations of standard random matrices (see \cite{liao2018dynamics} for related methods). This is the content of propositions \ref{eq:cauchy_representation} and \ref{eq:cauchy_representation_train}. With a natural concentration hypothesis in the high-dimensional regime,  it remains to carry out averages over the static traces involving random matrices. This is resolved using traces of sub-blocks from the inverse of a larger nontrivial block-matrix, a so-called \emph{linear pencil}. To the best of our knowledge linear pencils have been introduced in the machine learning community only recently in \cite{adlam2020neural}. This theory is developed in the context of random matrix theory in \cite{spectra, 8180454} and \cite{helton2018applications} using operator valued free-probability. A side-contribution in the SM is also an independent (but non-rigorous) derivation of a basic result of this theory (a fixed point  equation) using a replica symmetric calculation from statistical physics. The non-linearity of the activation function is addressed using the gaussian equivalence principle \cite{pennington2017nonlinear, 10.1214/19-ECP262, adlam2020neural}. 
Finally, our analysis is not entirely mathematically controlled mainly due to the concentration hypothesis in Sect. \ref{subsec:highdim} but comparison with simulations (see Fig. \ref{fig:tripledescent} and SM) confirm that the analytical results are exact. 

In the conclusion we briefly discuss possible extensions of the present analysis and open problems among which is the comparison with a dynamical mean-field theory approach.

%% file: main-theory.tex
\section{Random feature model} % Model setup, Results and insights (puis subsections)
%In this section, we setup the framework in which we will derive the dynamics of the gradient flow. 
%We first describe a learning task through a linear generative model along with a shallow neural network, define training and test errors, and 

\subsection{Model description}\label{subsec:modeldescription}
%\paragraph{Learning task:} 

\paragraph{Generative model and neural network:}
We consider the problem of learning a linear function $f_d(x) = d^{-\frac12}\beta^T x$ with $x, \beta\in \mathbb{R}^d$ column vectors. The vector $x$ is interpreted as a random input and $\beta$ as a random {\it hidden} vector; both with distribution $\Normal(0,I_d)$, $I_d$ the $d\times d$ identity matrix. We assume having access to the hidden function through the noisy outputs $y = f_d(x) + \epsilon$ with additive gaussian noise $\epsilon \sim \Normal(0,s^2)$, $s\in\mathbb{R}_+$. 
We suppose that we have $n$ data-points $(x_i, y_i)_{1 \leq i \leq n}$. This data can be represented as the $n\times d$ matrix $X \in \mathbb{R}^{n\times d}$ where $x_i^T$ is the $i$-th row of $X$, and the column vector vector $Y \in \Real^n$ with $i$-th entry $y_i$. Therefore, we have the matrix notation $Y = d^{-\frac12} X \beta + \xi$ where $\xi \sim \Normal(0,s^2 I_n)$ and $I_n$ the $n\times n$ identity matrix.

%\paragraph{Model structure:} 
We learn the data with a shallow $2$-layer neural network. There are $N$ hidden neurons with weight (column) vectors $\theta_i\in \mathbb{R}^d$, $i=1,\cdots, N$ each connected to the $d$ input neurons. Out of these we form the matrix (of the first layer connecting input and hidden neurons) 
$\Theta\in \mathbb{R}^{N\times d}$ where $\theta_i^T$ is the $i$-th row of $\Theta$ . Its entries are assumed independent and sampled through a standard gaussian distribution $\Normal(0,1)$; they are not learned but fixed once for all. The data-points in $X$ are applied linearly to the parameters $\Theta$, and the output $Z \in \mathbb{R}^{n\times N}$ of the first layer is the {\it pointwise application} of an activation function $\sigma: \mathbb{R} \to \mathbb{R}$, $Z = \sigma(d^{-\frac12} X \Theta^T)$. We use the notation $z_i^T$ to express the $i$-th row of $Z$. The second layer consists in a weight (column) vector $a_t \in \Real^N$ to be learned, indexed by time $t\geq 0$, with components initially sampled at $t=0$ i.i.d 
$\Normal(0,r^2)$, $r \in \Real_+$. 
%Thus $\norm{a_0} \to r$ when $N\to +\infty$ with probability tending to one. 
The prediction vector is expressed as $\hat Y_t = N^{-\frac12} Z a_t$.

We assume that the activation function belongs to $L^2(e^{-\frac{x^2}{2}} dx)$ with inner product denoted $\langle\, , \, \rangle$. It can be expanded on the basis of Hermite polynomials, so $\sigma \in \text{Span} \left( (H_{e_k})_{k \geq 0} \right)$, where
$H_{e_k}(x) = (-1)^k e^{\frac{x^2}{2}} \frac{d^k}{dx^k} e^{-\frac{x^2}{2}}$ (so $H_{e_0}(x)=1$, $H_{e_1}(x)=x$, $H_{e_2}(x)=x^2-1$, $H_{e_3}(x)=x^3-3x$, ...).
Furthermore we take $\sigma$ centered with $\langle \sigma, H_{e_0} \rangle = 0$, and set $\mu = \langle \sigma, H_{e_1} \rangle$, $\nu^2 = \langle \sigma, \sigma \rangle - \mu^2$. For instance, $\sigma = \text{id}$ has $(\mu,\nu) =  (1, 0)$ while $\sigma = \text{Relu} - \frac{1}{\sqrt{2 \pi}}$ has $(\mu,\nu) =  (\frac{1}{2}, \frac12 (1 - \frac{2}{\pi})^{1/2} ) \simeq (0.5, 0.3)$.
%define precisely smooth enough
Finally, we recall that we are interested in the high dimensional regime where the parameters $N,n,d$ tend to infinity 
with the ratios $\frac{N}{d} \to \psi$ and $\frac{n}{d}\to \phi$.

\paragraph{Training and test errors:} For a new input $x_0 \in \Real^d$, we define the predictor $\hat y_t(x_0)=\frac{1}{\sqrt N} z(x_0)^T a_t$ where $z(x_0) = \sigma(\frac{1}{\sqrt d} \Theta x_0 )$. We will further define the standard training and test errors with a 
penalization term $\lambda > 0$ and the quadratic loss:
\begin{equation}\label{eq:MSE}
\MSEtrain (a) = \frac{1}{n} \norm{Y - \hat Y}^2 + \frac{\lambda}{N} \norm{a}^2, \quad  \MSEtest (a) = \mean_{x_0 \sim \Normal(0,1)}{\left[(y(x_0)-\hat y(x_0))^2\right]}
\end{equation}
Note that because of the $\lambda$-penalization term, in this context, the training error can be above the test error in some configurations of parameters. Also, we will slightly abuse this notation throughout the paper by using $\MSEtrain_t, \MSEtest_t$ to designate $\MSEtrain (a_t), \MSEtest(a_t)$.

\paragraph{Gradient flow:} Minimizing the training error of this shallow-network is equivalent to a standard Tikhonov regularization problem with a design matrix $Z$ for which 
%provided invertibility (which is the case when $\lambda >0$) 
the optimal weights are given by $a_\infty = (\frac{Z^TZ}{N} + \frac{n}{N} \lambda I_N)^{-1} \frac{Z^T}{\sqrt N} Y$. The errors generated by the predictors with weights $a_\infty$ have been analytically calculated in the high-dimensional regime in \cite{mei2020generalization} and further analyzed in \cite{dascoli2020triple}. Here we study the {\it whole time evolution} of the gradient flow and thus introduce an additional time dimension in our model. Of course as $t\to+\infty$ one recovers the errors generated by the predictors with weights $a_\infty$. The output vector $a_t$ is updated through the ordinary differential equation $\frac{d a_t}{d t} = -\eta \nabla_a \MSEtrain (a_t)$ with a fixed learning rate parameter $\eta>0$. As $\eta$ can be absorbed in the time parameter, from now on we consider without loss of generality that $\eta = \frac{n}{2}$. Setting $\delta = \lambda \frac{n}{N}$, we find that the gradient flow for $a_t$ is a first order linear matrix differential equation,
\begin{equation}\label{eq:d_at}
    \frac{d a_t}{dt} = - \left(\frac{Z^T Z }{N} + \delta I_N\right) a_t + \frac{Z^T Y}{\sqrt N} .
\end{equation}
Recall the initial condition $a_0$ is a vector with i.i.d $\mathcal{N}(0, r^2)$ components.

\subsection{Cauchy integral representations of the training and test errors}
An important step of our analysis is the representation of  $\MSEtrain$ and $\MSEtest$ in terms of Cauchy contour integrals in the complex plane.
%Our method consists in deriving precise analytical results for $\MSEtrain$ and $\MSEtest$ when $N,n,d$ are large enough. 
To this end we decompose both errors in elementary contributions and derive contour integrals for each of them. Details are found in section  \ref{sec:sketch-proof} and the  SM. 
%appendices \ref{app:test-error-substitution}, \ref{app:cauchy-integral-repr}. 

We begin with the test error which is more complicated. We have 
\begin{align}\label{eq:test-error-decomposition}
\MSEtest_t  = 1+ s^2 -2\mu g(t) + \mu^2 h(t) + \nu^2 l(t) + o_d(1)
\end{align}
where $\lim_{d\to +\infty} o_d(1) = 0$ with high  probability,  and 
$g(t) = \frac{\beta^T}{\sqrt d} \frac{\Theta^T}{\sqrt d}  \frac{a_t}{\sqrt N}$, $h(t) = \Vert\frac{\Theta^T}{\sqrt d} \frac{a_t}{\sqrt N}\Vert^2$, and $l(t) = \Vert\frac{a_t}{\sqrt N}\Vert^2$.
To describe Cauchy's integral representation of the elementary functions $g$, $h$, $l$ we introduce the resolvent $R(z) = (\frac{Z^TZ}{N} - z I_N)^{-1}$ for all $z\in \Complex \setminus \Sp (\frac{Z^TZ}{N}) $. 

\begin{proposition}[Test error]\label{eq:cauchy_representation}
Let $\opR_z$ be the functional acting on holomorphic functions $f : \Complex \setminus \Sp(\frac{Z^TZ}{N}) \rightarrow \Complex$ as $\opR_z \{ f(z)\} = - \oint_{\Gamma} \frac{\dd z}{2\pi i} f(z)$ over a contour $\Gamma$ encircling the spectrum $\Sp (\frac{Z^TZ}{N})$ in the counterclockwise direction. Similarly, let $\opR_{x,y}$ be the functional acting on two-variable holomorphic functions $f : (\Complex \setminus \Sp (\frac{Z^TZ}{N}))^2 \rightarrow \Complex$ as $\opR_{x,y} \{ f(x, y)\} = \oint_{\Gamma}\oint_{\Gamma} \frac{\dd x}{2\pi i}\frac{\dd y}{2\pi i} f(x,y)$.
Let $G_t(z) = \frac{\beta^T}{\sqrt d} \frac{\Theta^T}{\sqrt d} R(z) \frac{a_t}{\sqrt N}$ and $K(z) = \frac{\beta^T}{\sqrt d} \frac{\Theta^T}{\sqrt d} R(z) \frac{Z^T Y}{N}$. We have for all $t\geq 0$
\begin{equation}\label{eq:repr_g} 
g(t) = \opR_z \left\{  e^{-t(z+\delta)} G_0(z) +  \frac{1 - e^{-t(z+\delta)}}{z+\delta} K(z)  \right\} .
\end{equation}
Let $L_t(z) = \frac{a_t^T}{\sqrt N} R(z) \frac{a_t}{\sqrt N}$ and $U_t(z) = \frac{Y^T Z}{N} R(z) \frac{a_t}{\sqrt N}$ and $V(z) = \frac{Y^T Z}{N} R(z) \frac{Z^T Y}{N}$. For all $t\geq 0$
    \begin{equation}
        l(t) \! = \! \opR_z \! \left\{ e^{-2t(z+\delta)} L_0(z)
        \! + \! 2 e^{-t(z+\delta)} \left( \frac{1 - e^{-t(z+\delta)} }{\delta +z} \right) U_0(z) 
        \! + \! \left(\frac{1-e^{-t (\delta +z)}}{\delta +z} \right)^2V(z)
        \right\} \! .
    \end{equation}
    Let $H_t(x,y) = \frac{a_t^T}{\sqrt N} R(x) \frac{\Theta \Theta^T}{d} R(y) \frac{a_t}{\sqrt N}$, $Q_t(x,y) = \frac{a_t^T}{\sqrt N} R(x) \frac{\Theta \Theta^T}{d} R(y) \frac{Z^T Y}{N}$ and $W(x,y) = \frac{Y^T Z}{N} R(x) \frac{\Theta \Theta^T}{d} R(y) \frac{Z^T Y}{N}$. For all $t\leq 0$
    \begin{align}\label{eq:repr_h}
            h(t) & =  
                \opR_{x,y} \left\{ 
                e^{-t (2 \delta +x+y)} \left(
                \frac{e^{ t (\delta + y) }- 1}{\delta + y} Q_0(x,y) +
                \frac{e^{ t (\delta + x) }- 1}{\delta + x} Q_0(y,x)
                \right) \right\}
                \nonumber \\ &
                + \opR_{x,y} \left\{ 
                    \frac{1-e^{-t(x+\delta)}}{x+\delta}
                    \frac{1-e^{-t(y+\delta)}}{y+\delta}
                 W(x,y) \right\}
                 + \opR_{x,y} \left\{ e^{-t(x+y+2\delta)} H_0(x,y) 
            \right\} .
    \end{align}
\end{proposition}

A similar but much simpler representation holds for the training error.

\begin{proposition}[Training error]\label{eq:cauchy_representation_train}
With the same definitions than in proposition \ref{eq:cauchy_representation} we have 
\begin{align}\label{eq:cauchy_train}
        \MSEtrain_t \! = \! \frac{\norm{Y}^2}{n} \! + \! \frac{1}{c} \opR_z  \!
        \left\{ (z+\delta) e^{-2t(z+\delta)} L_0(z)
       \!  - \! 2 e^{-2t(z+\delta)} U_0(z) \! - \! \frac{1 - e^{-2 t (\delta +z)}}{\delta +z} V(z) \!
        \right\}.
\end{align}
\end{proposition}

\subsection{High-dimensional framework}\label{subsec:highdim}

The Cauchy integral representation involves a set of one-variable functions $\mathcal S_1 = \{G_0,K,L_0,U_0,V\}: \Complex \setminus \text{Sp}(\frac{Z^TZ}{N}) \to \Complex$ and a set of two-variable functions $\mathcal S_2 = \{H_0,W,Q_0\}: \Complex \setminus \text{Sp}(\frac{Z^TZ}{N}))^2 \to \Complex$ so that $g$, $h$, $l$ and thus also $\MSEtest$ and $\MSEtrain$ are actually functions of $(t;\mathcal S_1,\mathcal S_2)$. Thus we can write for instance: $\MSEtest(a_t)=\MSEtest(t;\mathcal S_1,\mathcal S_2)$.
%We restrict exclusively our work on recovering the full path of the time evolution of the training and test error, it is clear that these quantities depend on underlying matrices and vectors $X,Y,\Theta,\xi,\beta,a_0$ included $\mathcal S_1, \mathcal S_1$ and whose sizes depend on $N,n,d$. 
%It is clear that the time evolution of the training and test error, it is clear that these quantities depend on underlying matrices and vectors $X,Y,\Theta,\xi,\beta,a_0$ and $\mathcal S_1, \mathcal S_2$, whose sizes depend on $N,n,d$. 
We simplify the problem by considering the high-dimensional regime where $N, n, d\to \infty$ with ratios $\frac{N}{d} \to \psi$, $\frac{n}{d}\to \phi$ tending to fixed values of order one. 
In this regime we expect that the functions in $\mathcal S_1$ and $\mathcal S_2$ concentrate and can therefore be replaced by their averages over randomness. These averages can be carried out using recent progress in random matrix theory \cite{spectra}, \cite{8180454}, and we are able to compute pointwise asymptotic values of the functions in $\mathcal S_1$, $\mathcal S_2$, and eventually substitute them in the Cauchy integral representations for the training and test error. In general, rigorously showing concentration of the various functions involved is not easy and we will make the following assumptions:

\begin{assumptions}\label{assumptions:highdim}
In the high dimensional limit with $d, N, n \to \infty$ and  $\frac{N}{d}\to\psi$, $\frac{n}{d} \to \phi$: 
%the following assumptions holds:
\begin{enumerate}
        \item The random functions in $\mathcal S_1$, $\mathcal S_2$ are assumed to concentrate. 
        We let $\bar{\mathcal S}_1 = \{\bar G_0,\bar K,\bar L_0,\bar U_0,\bar V\}$ and $\bar{\mathcal S}_2 = \{\bar H_0,\bar W,\bar Q_0\}$ be the pointwise limit of the functions.
        \item There exists a bounded subset $\mathcal C \subset \Real^+$ such that the functions in $\bar{\mathcal S}_1$ and $\bar{\mathcal S}_2$ are holomorphic on $\Complex \setminus \mathcal C$ and $(\Complex \setminus \mathcal C)^2$ respectively
        \item The gaussian equivalence principle (see sect. \ref{def:gausseq}) can be applied to the limiting quantities.
\end{enumerate}
\end{assumptions}

It is common that the closure of the spectrum of suitably normalized random matrices  concentrates on a deterministic set. Thus the bounded set $\mathcal C$ can be understood as the limit of the finite interval $[0, \lim_d \max\text{Sp}(\frac{Z^TZ}{N})]$.  In the sequel we will distinguish the theoretical high-dimensional regime from the finite dimensional regime using the upper-bar notation. 

\begin{definition}[High-dimensional framework]\label{def:framework}
Under the assumptions \ref{assumptions:highdim}, we define the theoretical test error $ \barMSEtest_t = \MSEtest(t; \bar{\mathcal S}_1, \bar{\mathcal S}_2)$ and the theoretical training error $\barMSEtrain_t = \MSEtrain(t; \bar{\mathcal S}_1, \bar{\mathcal S}_2)$
\end{definition}

We conjecture that  $\lim_d \MSEtrain_t = \barMSEtrain_t$ and $\lim_d \MSEtest_t = \barMSEtest_t$ at all time $t \in \mathbb R$. We verify that this conjecture stands experimentally for sufficiently large $d$ on different configurations (see additional figures in the SM). 
This also lends experimental support on the assumption \ref{assumptions:highdim}.
Furthermore we conjecture that the $d\to +\infty$ and $t \to \infty$ limits commute, namely $\lim_d \lim_t \MSEtrain_t = \lim_t \barMSEtrain_t$ and $\lim_d \lim_t \MSEtest_t = \lim_t \barMSEtest_t$.

%% file: main-result.tex
\subsection{Main results}

In this section we provide the main results of this work:  analytical formulas tracking the test and training errors during gradient flow of the random feature model for all times in the high-dimensional theoretical framework.

\begin{result}\label{th:main}
    Under the assumption \ref{assumptions:highdim}, the theoretical test and training errors of definition \ref{def:framework} are given for all times $t\geq 0$ by the formulas
    \begin{equation}
        \barMSEtest_t = 
        1+s^2
        - 2 \mu \bar g(t) + \mu^2 \bar h(t) + \nu^2 \bar l(t) ,
    \end{equation}
    \begin{equation}
        \barMSEtrain_t = 1 + s^2 + \frac{1}{c} \int_{\Real} \left[
        (\delta+\omega) e^{-2t(\omega+\delta)} \rho_{\bar L_0}(\omega)
        - \frac{1-e^{-2 t (\delta + \omega)}}{\delta + \omega} \rho_{\bar V}(\omega)
        \right] \dd \omega ,
    \end{equation}
    with $c = \frac{\phi}{\psi}$, $\delta = c\lambda$, and the functions $\bar g$, $\bar h$, $\bar l$ given by
    \begin{equation}
        \bar g(t) = \int_{\Real} 
        \frac{1-e^{-t (\omega + \delta)}}{\omega + \delta}
        \rho_{\bar K}(\omega) \dd \omega ,
    \end{equation}
    \begin{equation}
        \bar l(t) = \int_{\Real} \left[
        e^{-2t(\omega + \delta)} \rho_{\bar L_0}(\omega)
        +
        \left( \frac{1-e^{-t (\omega + \delta)}}{\omega + \delta} \right)^2
        \rho_{\bar V}(\omega) \right] \dd \omega ,
    \end{equation}
    \begin{equation}
        \bar h(t) = \iint_{\Real^2} \left[
        e^{-t(u+v+2\delta)} \rho_{\bar H_0}(u,v) 
        +
        \frac{1-e^{-t (u + \delta)}}{u + \delta}
        \frac{1-e^{-t (v + \delta)}}{v + \delta}
        \rho_{\bar W}(u,v) 
        \right] \dd u \dd v ,
    \end{equation}
    where the measures $\rho_{\bar K}, \rho_{\bar L_0}, \rho_{\bar V}, \rho_{\bar H_0}, \rho_{\bar W}$ (are possibly signed) are characterized by their Stieltjes transforms given by $\bar K, \bar L_0, \bar V, \bar H_0, \bar W$
    \begin{align}
    \begin{cases}
    \bar K(x) = t_1^x, \quad
    \bar L_0(x) = r^2  g_1^x, \quad
    \bar V(x) = s^2 \left( 1 + x g_1^x \right) + \left(c-h_4^x\right), \\
    \bar H_0(x,y) = r^2 q_1, \quad
    \bar W(x,y) = s^2 cq_4 + q_2
    \end{cases}
    \end{align}
    where for  each $x,y \in \mathbb{C}^+$ (the upper half complex plane) $g_1^x, h_4^x, t_1^x, g_1^y, h_4^y, t_1^y$ and $q_1, q_2, q_4, q_5$ (which depend symmetrically on $(x,y)$, e.g., $q_1 = q_1^{x,y} = q_1^{y,x}$) are solutions of a purely algebraic system of equations (see SM for the criterion to select the relevant solution)
    \begin{align*}
        \left\{ \begin{array}{l}
            0 = \mu \psi g_1^x h_4^x - t_1^x\\
            0 = \mu \psi g_1^y h_4^y - t_1^y\\
            0 = (c-1-x g_1^x) \left(c - \mu^{2} \phi g^{x}_{1} h^{x}_{4} \right) - c h^{x}_{4} \\
            0 = (c-1-y g_1^y) \left(c - \mu^{2} \phi g^{y}_{1} h^{y}_{4} \right) - c h^{y}_{4}\\
            0 = 1 - g^{x}_{1} \left(\mu^{2}  h^{x}_{4} + (c-1-x g_1^x) \nu^{2} -  x\right)\\
            0 = 1 - g^{y}_{1} \left(\mu^{2}  h^{y}_{4} + (c-1-y g_1^y) \nu^{2} - y\right)\\
            0 = - \mu^{2} g^{y}_{1} q_{2} + \mu^{2} h^{x}_{4} q_{1} + \mu g^{y}_{1} t^{x}_{1} + \mu g^{y}_{1} t^{y}_{1} - c g^{y}_{1} q_{4}\nu^{2} - g^{y}_{1} - q_{1} x + q_{1}\nu^{2} \left(c - g^{x}_{1} x - 1\right)\\
            0 = \mu \left(\phi - \psi g^{x}_{1} x - \psi\right) \left(- \mu g^{x}_{1} q_{2} + \mu h^{y}_{4} q_{1} + g^{x}_{1} t^{y}_{1}\right) + c q_{4} (1-\mu t_1^y) - q_{2}\\
            0 = - \mu^{2} \phi g^{x}_{1} (1-\mu t_1^x) q_{4} + \mu^{2} q_{5} \left(c - g^{y}_{1} y - 1\right) - \nu^{2} \phi g^{x}_{1} q_{4} - \phi q_{4} + q_{1}\nu^{2} \left(\phi - \psi g^{y}_{1} y - \psi\right)\\
            0 = \psi (\mu^{2} \phi g^{x}_{1} g^{y}_{1} q_{4} + \psi g^{x}_{1} g^{y}_{1}  +  q_{1}) (1-\mu t_1^y)  - \mu^{2} \psi g^{x}_{1} q_{5} \left( c - g^{x}_{1} x - 1 \right)  - q_{5}
        \end{array}\right.
    \end{align*}
\end{result}

%Therefore, the analytical formulas for the functions $\rho_K, \rho_{L_0}, \rho_V, \rho_{H_0}, \rho_W$ allow to fully compute numerically the full trajectory of the test error at any time $t$. As conjectured, this result matches quite well the result of a large RF model where $d$ is taken infinitely huge. The latter model quickly becomes intractable to simulate with a usual gradient descent while the formulas allow to quickly predict the evolution at any time - with the compromise of not recovering any vector weight $a_t$. Therefore, we expect this method can be an impactful way to adjust the training resources and parameter size prior to running a model.

We can also deduce the limiting training error and test errors in the infinite time limit: 
%This is known to be the minimum least squares estimator in the $d$-dimensional regime.
\begin{result}\label{result-infinite-time}
    In the limit $t\to \infty$ we find:
    \begin{align*}
        \lim_{t\to \infty} \barMSEtest_{t}
        \! = \! 1+s^2 - 2 \mu \bar K(-\delta) + \mu^2 \bar W(-\delta,-\delta)
        + \nu^2 \frac{\dd \bar V}{\dd x} (-\delta), 
        \,\,
        \lim_{t\to \infty} \barMSEtrain_{t}
        \! = \! 1+s^2 - \frac{1}{c} \bar V(-\delta)
    \end{align*}
\end{result}
Interestingly, in the limit $t\to \infty$, the expressions become simpler and completely algebraic in the sense that we do not need to compute integrals (or double-integrals) over the supports of the eigenvalue distributions. It is not obvious to see on the analytical expressions  that the result is the same as the algebraic expressions obtained in \cite{mei2020generalization} but Fig. \ref{fig:tripledescent} shows an excellent match with simulation experiments. We note here that checking that two sets of complicated algebraic equations are equivalent is in general a non-trivial problem of computational algebraic geometry \cite{GrobnerBasesBook}.

%% file: main-insights.tex
\subsection{Insights and illustrations of results}

\begin{figure}[h!]
  \centering
  \subfloat{
    \includegraphics[width=0.4\textwidth]{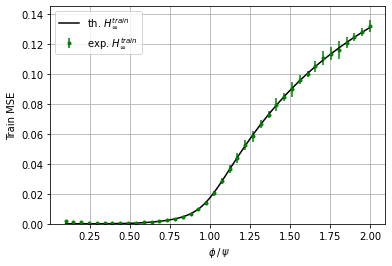}
  }
  \subfloat{
    \includegraphics[width=0.4\textwidth]{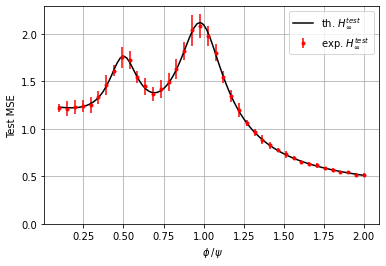}
  }
  \caption{\small {\it Large time limit.}
    Analytical training error and test error profile with parameters
    $(\mu,\nu,\psi,r,s,\lambda) = (10,1,2,1,0.5,0.01)$ compared with experimental least squares MSE with $40$ data-points with $d = 5000$ (average of 10 instances with confidence bar at $2\sigma$)
  }
  \label{fig:tripledescent}
\end{figure}

The set of analytical formulas allows to compute numerically the measures $\rho_K, \rho_{L_0}, \rho_V, \rho_{H_0}, \rho_W$ and in turn the full time evolution of the test and training errors. The result matches the simulation of a large random feature model where $d$ is taken large as can be seen on Figs. \ref{fig:tripledescent} for the infinite time limit (experimental check of result \ref{result-infinite-time})
and additional figures in the SM (experimental check of result \ref{th:main}). 
%Large models quickly become intractable to simulate with usual gradient descent, and the parameter space $(t,\mu,\nu,\psi,\phi,r,s,\lambda)$ is already quite large, thus the analytical formulas are a practical guide to predict the evolution at any time, and provide a guide to adjust the training resources and parameter sizes.
Below we illustrate numerical computations obtained with analytical formulas of result \ref{th:main} for various sets of parameters $(t,\mu,\nu,\psi,\phi,r,s,\lambda)$. For instance, we can freely choose two of these parameters and plot the generalization error in 3D as in Fig. \ref{fig:intro}, or as a heat-map in the following. We describe three important phenomena which are observed with our analysis.

\begin{figure}[h!]
  \centering
  \includegraphics[width=1.0\textwidth]{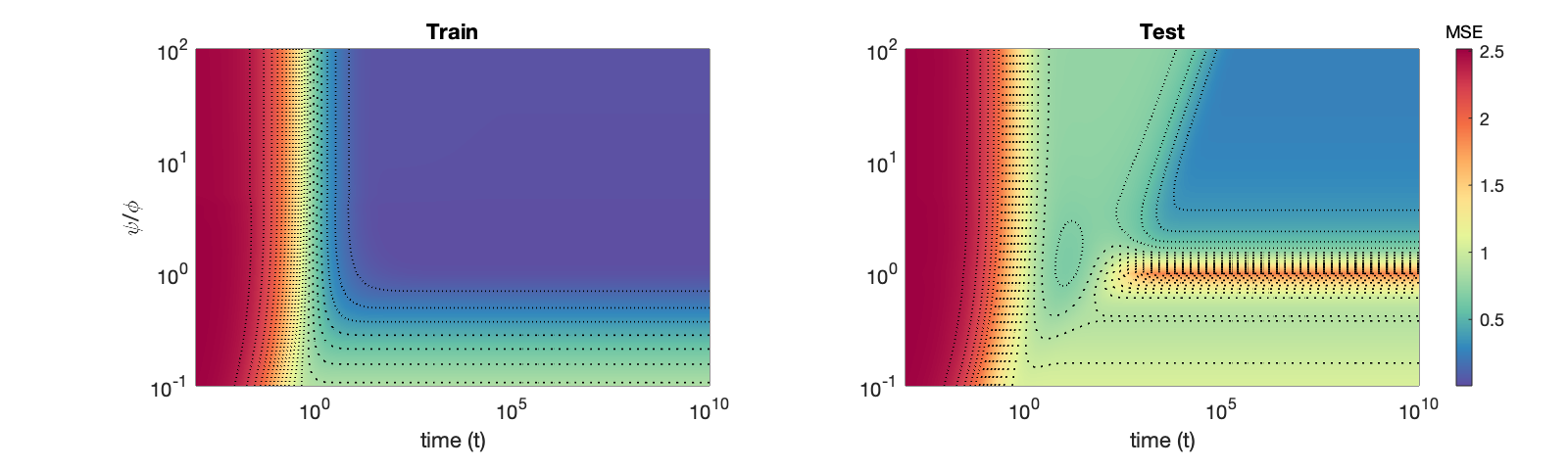}
  \caption{\small {\it Model-wise double descent.}
  Analytical training error and test error evolution with parameters
  $(\mu,\nu,\phi,r,s,\lambda) = (0.5,0.3,3,2.,0.4,0.001)$. Note that we vary the number of model parameters ($\psi$).
  }
  \label{fig:time_double_descent0}
\end{figure}

\emph{Double descent and early-stopping benefits:} while \cite{mei2020generalization} mostly analyze the minimum least-squares estimator of the random feature model which displays the double-descent at $\psi=\phi$, we are predicting the whole time evolution of the gradient flow as in Fig. \ref{fig:time_double_descent0}. We clearly observe the double-descent curve at $t=10^{10}$ for $\psi=\phi$; but we now notice that if we stop the training earlier, say at times $1<t<10$, the generalization error performs better than the minimum least squares estimator. Actually, in the time interval $t\in(1,10)$ for $\psi \approx \phi$ the test error even has a {\it dip or minimum} just before the spike develops. We also notice a two-steps descent structure with the test error which is non-existent in the training error and materializes long after the training error has stabilized in the overparameterized regime $\psi \gg \phi$. This is also reminiscent but not entirely similar to the abrupt \textit{grokking} phenomenon described in \cite{power2021grokking}.

\begin{figure}[h!]
  \centering
  \includegraphics[width=1.0\textwidth]{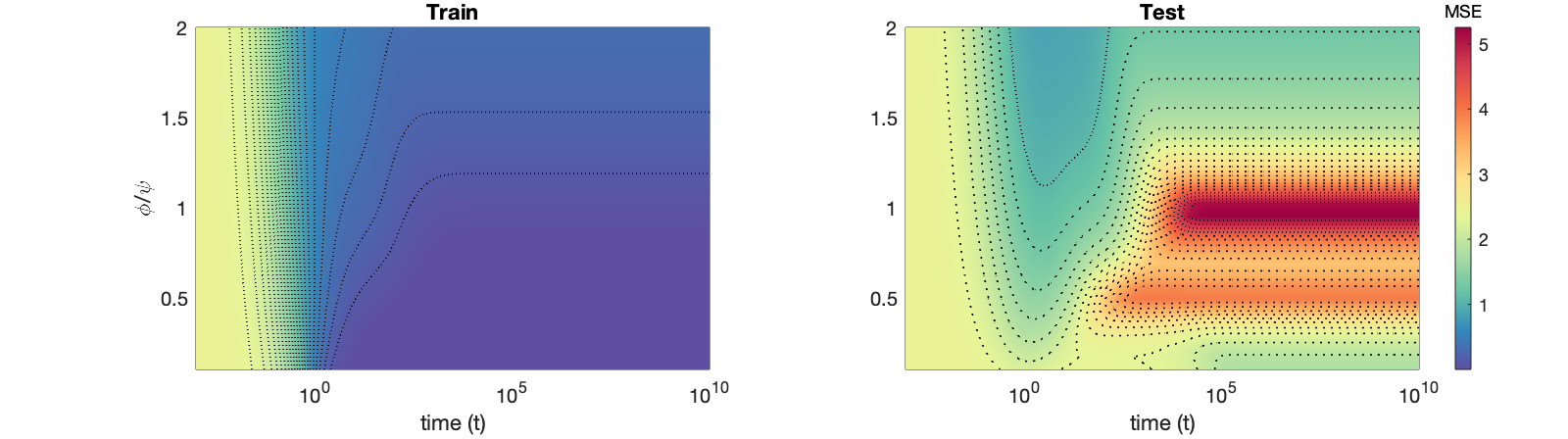}
  \caption{\small {\it Sample-wise descents.}
  Analytical training error and test error evolution with parameters
  $(\mu,\nu,\psi,r,s,\lambda) = (0.9,0.1,2,1,0.8,0.0001)$. Note that we vary the number of samples ($\phi$).
  }
  \label{fig:time-tripledescent}
\end{figure}

\emph{Triple descent:} We can observe a triple descent phenomenon materialized by two spikes as seen in Fig. \ref{fig:tripledescent} at $t=\infty$ (we also check that the theoretical result matches very well the empirical prediction of the minimum least squares estimator both for training and test errors). This triple descent phenomenon is already contained in the formulas of 
\cite{mei2020generalization} (although not discussed in this reference) and has been analyzed in detail in \cite{dascoli2020triple}.
The test error contains a so-called {\it linear spike} for $\phi = 1$ ($n=d$) and a  {\it non-linear} spike for $\psi = \phi$ ($N=n$). The two spikes are often not seen together as this requires certain conditions to be met, and they tend to materialize together for specific values $\mu$, $\nu$ of the activation function where $\mu \gg \nu$. Here we further observe the evolution through time of the triple descent and the two spikes and how they develop in Fig. \ref{fig:time-tripledescent}. There, we notice that the linear-spike seems to appear {\it earlier} than the non-linear one. 

\begin{figure}[h!]
  \centering
  {
    \includegraphics[width=1.0\textwidth]{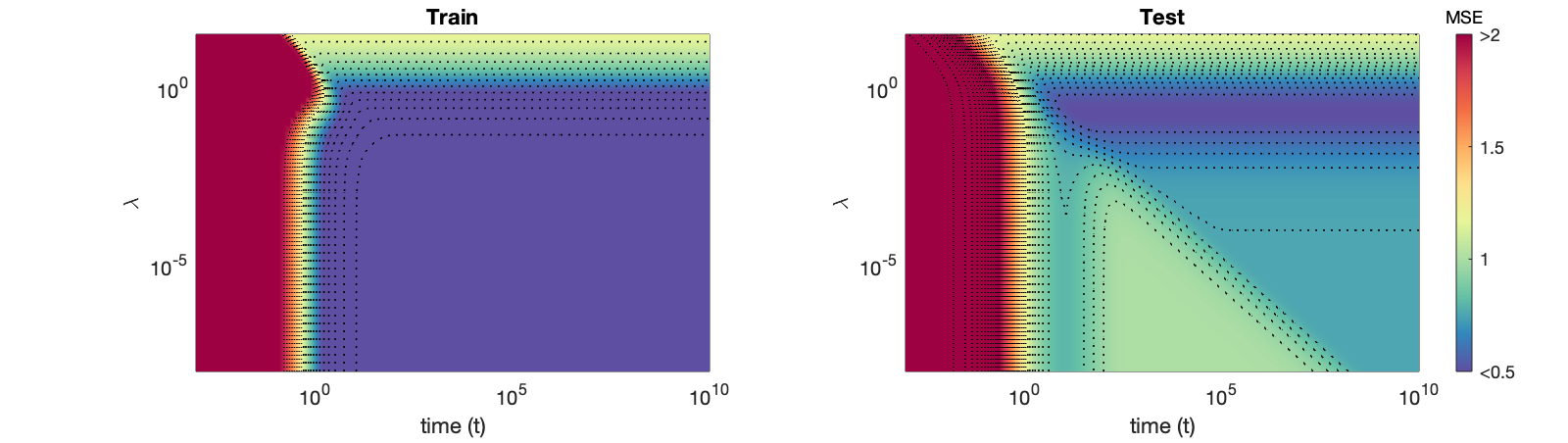}
  }
  \caption{\small {\it Epoch-wise descent structures.}
    Analytical test error evolution with respect to different values of $\lambda$
    $(\mu,\nu,\psi,\phi,r,s) = (0.5,0.3,6,3,2.0,0.5)$. Here the ratio of number of parameters and samples is fixed. 
  }
  \label{fig:time_double_descent}
\end{figure}

\emph{Epoch-wise descent structures:}  Important phenomena that we uncover here are two time-wise "descent structures". (i) As can be seen in Fig. \ref{fig:time_double_descent0}, the test error develops a {\it double plateau} structure at widely different time scales in the over-parameterized regime ($\psi \gg \phi$) while there seems to be only one time scale for the training error. This kind of double plateau descent is different from the "usual" double-descent. (ii) Moreover, on Fig. \ref{fig:time_double_descent} for well chosen parameters (in particular for noises with $s$ and $r$ "larger" and $\psi = 2 \phi$), we can also observe an {\it elongated bump} (rather than a thin spike) for small $\lambda$'s. 
Notice the logarithmic time-scale which clearly shows that here we need to wait exponentially longer to attain the "second descent" after the bump. 
This is very reminiscent of the epoch-wise double descent described in \cite{nakkiran2019deep} for deep networks (which happens on similar time scales). 
%This phenomenon shows that we can't always safely rely only on a reversing of the test error descent as an early stopping indicator in our training algorithm. 


%\begin{figure}[h!]
%  \centering
%  \subfloat[$s=0.4$]{
%    \includegraphics[width=0.45\textwidth]{images/s_MSE.png}
%  }
%  \subfloat[$r=2$]{
%    \includegraphics[width=0.45\textwidth]{images/r_MSE.png}
%  }
%  \caption{ 
%    Analytical test error evolution with respect to different values of $r,s$
% $(\mu,\nu,\psi,\phi,\lambda) = (0.5,0.3,6,3,0.0001)$ 
%  }
%  \label{fig:time_double_descent2}
%\end{figure}

%% file: main-sketch-proof.tex
\section{Sketch of proofs and analytical derivations}\label{sec:sketch-proof}

The analysis is threefold. Firstly, we decompose the training and test errors in elementary terms and establish Cauchy's integral representation for each of them, as provided in proposition \ref{eq:cauchy_representation}. A crucial advantage of this form is that it dissociates a scalar time-wise component and static matrix terms. Secondly, we switch to the high-dimensional framework where the matrix terms are substituted by their limit using the gaussian equivalence principle. Thirdly, we can compute the expectations of matrix terms thanks to a random matrix technique based on linear pencils. In this section we only sketch the main ideas for each step and provide details in the supplementary material.
 
\subsection{Cauchy's integral representation}\label{sec:cauchyproof_g}
We sketch the derivation for the test error and leave details to appendices. The derivation for the training error is entirely found in the SM.
Expanding the square in Equ. \eqref{eq:MSE} and carrying  out averages we find Equ. \eqref{eq:test-error-decomposition} for $\MSEtest_t$ with 
$g(t) = \frac{\beta^T}{\sqrt d} \frac{\Theta^T}{\sqrt d}  \frac{a_t}{\sqrt N}$, $h(t) = \Vert\frac{\Theta^T}{\sqrt d} \frac{a_t}{\sqrt N}\Vert^2$, and $l(t) = \Vert\frac{a_t}{\sqrt N}\Vert^2$ (see SM for this derivation).

We show how to derive the Cauchy integral representation for $g(t)$. For $h(t)$, $l(t)$ the steps are 
similar and found in SM. Let us consider the function $(t,z) \mapsto G_t(z)$ as in \ref{eq:cauchy_representation}.
Then we have the relation $g(t) = \frac{-1}{2i \pi} \oint_{\Gamma} \dd z G_t(z)$ where $\Gamma$ is a loop in $\Complex$ enclosing the spectrum of $\frac{Z^TZ}{N}$. This can easily be seen by decomposing the symmetric $\frac{Z^TZ}{N}$ in an orthonormal basis $v_1, \ldots, v_N$ with the eigenvalues $\lambda_1, \ldots, \lambda_N$: then we have $G_t(z) = \sum_{i=1}^N \frac{1}{\lambda_i-z} \left( \frac{\beta^T}{\sqrt d} \frac{\Theta^T}{\sqrt d} v_i v_i^T \frac{a_t}{\sqrt N} \right)$ and because $\lambda_i$ are all encircled by $\Gamma$, we find $- \oint_{\Gamma} \frac{\dd z}{2\pi i} G_t(z) = \sum_{i=1}^N \frac{\beta^T}{\sqrt d} \frac{\Theta^T}{\sqrt d} v_i v_i^T \frac{a_t}{\sqrt N} = g(t)$.
Now, the ODE derived for $a_t$ in \eqref{eq:d_at}, can be written slightly differently using the fact that $\frac{Z^TZ}{N} = R(z)^{-1} + zI$ for any $z$ outside $\text{Sp}(\frac{Z^TZ}{N})$. Namely, 
$ \frac{d a_t}{dt} = \frac{Z^T Y}{\sqrt N} - R(z)^{-1} a_t - (z + \delta) a_t $.
Then, we can derive an integro-differential equation for $G_t(z)$ involving $g(t)$ and $K(z)$:
\begin{equation}\label{eq:d_Gt}
    \frac{\partial_t G_t(z)}{\partial t} =
    K(z) - g(t) - (z+\delta) G_t(z)
\end{equation}
In the following, we let $\opL$ be the Laplace transform operator $(\opL f)(p) = \int_0^{+\infty} dt\, e^{-pt} f(t)$, $\Re p$ large enough. Note that the contour integral is performed over a compact set $\Gamma$ so for $\Re p$ large enough, by Fubini's theorem, the operations $\opL$ and $\opR_z$ commute. Applying $\opL$ to \eqref{eq:d_Gt} and rearranging terms we find for $\Re(p+z+\delta) \neq 0$:
\begin{equation}\label{eq:l_Gt}
    \opL G_p(z) = \frac{G_0(z)}{p+z+\delta} + \frac{K(z)}{p(p+z+\delta)} - \frac{\opL g(p)}{p+z+\delta}
\end{equation}
Now, we can always choose $\Gamma$ such that $-(p+\delta)$ is outside of the contour if we assume $\Re(p+\delta)>0$ (since $\min_i\lambda_i \geq 0$). Thus, applying $\opR_z$ to \eqref{eq:l_Gt} nullifies the last term because the pole is outside $\Gamma$, and using commutativity $\opR_z \opL G_p = \opL \opR_z G_p$,
\begin{align}
        \opR_z \opL G_p 
         = \opL \opR_z \left\{  e^{-t(z+\delta)} G_0(z) +  \frac{1 - e^{-t(z+\delta)}}{z+\delta} K(z)  \right\} = \opL g(p) .
\end{align}
%\begin{align}
%        \opR_z \opL G_p = \opL g(p) 
%        & = 
%        \opR_z \left\{ \opL \{ e^{-t(z+\delta)} \} G_0(z) \right\} + 
%        \opR_z \left\{ \opL \left\{ \frac{1 - e^{-t(z+\delta)}}{z+\delta}  \right\} K(z) \right\} 
%        \nonumber \\ &
%         % = \opR_z \opL \left\{  e^{-t(z+\delta)} G_0(z) +  \frac{1 - e^{-t(z+\delta)}}{z+\delta} K(z)  \right\}\\
%         = \opL \opR_z \left\{  e^{-t(z+\delta)} G_0(z) +  \frac{1 - e^{-t(z+\delta)}}{z+\delta} K(z)  \right\}
%\end{align}
Finally, using the inverse Laplace transform leads to \eqref{eq:repr_g}. 
%A similar method applied the other terms $h(t)$ and $l(t)$ is described further in the appendix.

\subsection{Gaussian equivalence principle}\label{def:gausseq}
The matrix terms must be estimated in the limit $d \to \infty$ with $\{\beta,a_0,\xi, \Theta, X\}$ all independently distributed. As per assumptions \ref{assumptions:highdim} all the matrix terms in $\mathcal S_1, \mathcal S_2$ are assumed to concentrate. So for instance we assume that the following limit exists $\bar K(z) \equiv \lim_{d \to \infty} K(z) = \lim_{d \to \infty} \mean_{\beta,\xi,\Theta,X} [K(z)]$. Using cyclicity of the trace we easily perform averages over $\beta, \xi$ to find
\begin{align}\label{eq:Klim}
    \bar K(z) = \lim_d \mean_{\beta, \Theta, X} \trace{ \frac{\Theta^T}{\sqrt d} R(z) \frac{Z^T X}{N} \frac{\beta \beta^T}{d} }
    = \lim_d \frac{1}{d} \mean_{\beta, \Theta, X} \trace{ \frac{\Theta^T}{\sqrt d} R(z) \frac{Z^T X}{N} } .
\end{align}
After these reductions, the expressions of all functions in $\bar{\mathcal S}_1, \bar{\mathcal S}_2$ essentially involve products of random matrices $\Theta$, $X$ and pointwise applications of the non-linear activation $\sigma$. This can be further reduced to simpler algebraic expressions using the {\it gaussian equivalence principle}. This principle states that: {\it there exists a standard gaussian random matrix $\Omega \in \mathbb{R}^{n\times N}$ independent of $\{X,\Theta\}$ such that in the infinite dimensional limit we can make the substitution $Z = \sigma \left( d^{-1/2} X \Theta^T \right) \longrightarrow \mu d^{-1/2} X \Theta^T + \nu \Omega$
in the expressions of all functions in $\bar{\mathcal S}_1, \bar{\mathcal S}_2$}.
%\begin{equation}\label{eq:gausseq}
%    Z = \sigma \left( \frac{1}{\sqrt d} X \Theta^T \right) 
%    \longrightarrow
%    \frac{\mu}{\sqrt d} X \Theta^T + \nu \Omega
%\end{equation}
This approach is quite general and is well described in \cite{adlam2020neural, adlam2019random} (and formerly in \cite{pennington2017nonlinear} and \cite{10.1214/19-ECP262}). Thus it remains to compute expectations of traces containing only products, and inverses of products and sums, of gaussian matrices.

\subsection{Expectations over random matrices using linear pencils}
We explain how to compute the limit of \eqref{eq:Klim} once the gaussian equivalence principle has been applied. A powerful approach is to design a so-called {\it linear pencil}. In the present context this is a suitable block-matrix containing gaussian random matrices and multiples of the identity matrix, for which full block-inversion gives back the products of terms in the traces that are being sought. This approach has been described in \cite{spectra, 8180454, mingo2017free}. We have found a suitable linear pencil which contains fortuitously {\it all} the terms required in $\bar{\mathcal S}_1, \bar{\mathcal S}_2$. It is described by the $13\times{13}$ {\it block-matrix} $M$, and pursuing with our example, we get for instance with the block $(7,12)$ that $\lim_d K(y) = \lim_d \frac{1}{d} {\rm Tr} [(M^{-1})^{(7,12)}]$
{\small
\begin{align*}
%\label{eq:Mxy}
    \begin{array}{c}
        M \!  = \! \!
        \left[ \!
            \begin{array}{ccccccccccccc}
                -xI & -\mu \frac{\Theta}{\sqrt d} & -I & 0 & 0 & 0 & \frac{\Theta}{\sqrt d} & 0 & 0 & 0 & 0 & 0 & 0\\ %\hline
                0 & I & 0 & \frac{X^T}{\sqrt N} & 0 & 0 & 0 & 0 & 0& 0 & 0 & 0 & 0\\
                0 & 0 & I & \nu \frac{\Omega^T}{\sqrt N} & 0& 0 & 0 & 0 & 0& 0 & 0 & 0 & 0\\
                0 & 0 & 0 & I & \frac{X}{\sqrt N} & \nu \frac{\Omega}{\sqrt N} & 0& 0 & 0 & 0& 0& 0  & 0\\
                \mu \frac{\Theta^T}{\sqrt d} & 0 & 0 & 0 & I & 0 & 0 & 0 & 0 & 0 & 0& 0 & 0\\ 
                I & 0 & 0 & 0 & 0 & I & 0 & 0 & 0 & 0 & 0& 0& 0 \\ %\hline \hline
                0 & 0 & 0 & 0 & 0 & 0 & I & 0 & 0 & 0& 0 & 0 & \frac{\Theta^T}{\sqrt d} \\ %\hline 
                0 & 0 & 0 & 0 & 0 & 0 & 0 & I & 0 & \nu \frac{\Omega^T}{\sqrt N} & 0 & 0 & 0\\ 
                0 & 0 & 0 & 0 & 0 & 0 & 0 & 0 & I & \frac{X^T}{\sqrt N} & 0 & 0 & 0\\ 
                0 & 0 & 0 & 0 & 0 & 0 & 0 & 0 & 0 & I & \nu \frac{\Omega}{\sqrt N} & \frac{X}{\sqrt N} & 0\\ 
                0 & 0 & 0 & 0 & 0 & 0 & 0 & 0 & 0 & 0 & I & 0 & -I\\
                0 & 0 & 0 & 0 & 0 & 0 & 0 & 0 & 0 & 0 & 0 & I & -\mu \frac{\Theta^T}{\sqrt d}\\ %\hline
                0 & 0 & 0 & 0 & 0 & 0 & 0 & I & \mu \frac{\Theta}{\sqrt d} & 0 & 0 & 0 & -y I\\ 
        \end{array}
        \!
        \right]
    \end{array}
\end{align*}
}
Next, the great advantage of the linear pencil is that (as described in \cite{spectra, 8180454, mingo2017free}) it allows to write a fixed point equation $F(G)=G$ for a "small" $13\times{13}$ matrix $G$ with {\it scalar} matrix elements. We also provide in the SM an independent derivation of the fixed point equations using the replica method (a technique from statistical physics \cite{edwards1976eigenvalue}).The components of $G$ are linked to the limiting traces of the blocks of $M^{-1}$ as in $[G]_{2,7} = \bar K(z)$. The action of $F$ can be completely described as an algebraic function leading to (a priori) $13 \times 13 =169$ equations over the matrix elements of $G$. The number of equations can be immediately reduced to $39$ because many elements vanish, and with the help of a computer algebra system the number of equations can be further brought down to $10$. We refer to the SM for all the details about the method.

%% file: main-conclusion.tex
\section{Conclusion} 

We believe that our analysis could be extended to study 
the learning of non-linear functions, the effect of multilayered structures, and potentially different layers such as convolutions, 
as long as they are not learned. A challenging task is to extend the present methods to learned multilayers. A further question is the application of our analysis in teacher-student scenarios with realistic datasets (See \cite{loureiro2021capturing,adlam2019random}).

Finally we wish to point out that a comparison of the approach of the present paper (and the similar but simpler one of \cite{pmlr-v134-bodin21a}) with the dynamical mean-field theory (DMFT) approach of statistical physics remains to be investigated. DMFT has a long history originating in studies of complex systems (turbulent fluids, spin glasses) where one eventually derives a set of complicated integro-differential equations
%, the so-called CSHCK equations  
for suitable correlation and response functions capturing the whole dynamics of the system (we refer to the recent book \cite{parisi2020theory} and references therein). This is a powerful formalism but the integral equations must usually be solved entirely numerically which itself is not a trivial task. For problems close to the present context (neural networks,generalized linear models, phase retrieval)  DMFT  has been developed 
% and 
in the recent works \cite{PhysRevLett.61.259, PhysRevE.98.062120, Biroli2018, NEURIPS2020_6c81c83c, Mignacco-et-al2021}. We think that comprehensively comparing this formalism with the present approach is an interesting open problem. It would be desirable to connect the DMFT equations to our closed form solutions for the training  and generalization errors expressed in terms of a set of algebraic equations of suitable Stieltjes transforms.

%% file: app-test-error.tex
\section{Test Error substitutions}\label{app:test-error-substitution}

The test error $\MSEtest_t$ in \eqref{eq:MSE} can be expanded into smaller terms
\begin{align}
        \MSEtest_t & = \mean_{x_0}[y(x_0)^2] - 2 \mean_{x_0} [y(x_0) \hat y_t(x_0)] + \mean_{x_0} [\hat y_t(x_0)^2]
        \nonumber \\ &
        = \mean_{x_0}[y(x_0)^2] - 2 \frac{\beta^T}{\sqrt d} \mean_{x_0} [x_0 z(x_0)^T] \frac{a_t}{\sqrt N} + \frac{a_t^T}{\sqrt N} \mean_{x_0} [z(x_0) z(x_0)^T] \frac{a_t}{\sqrt N} .
        \label{MSEtest-t}
\end{align}
The random noise $\epsilon$ from $y(x_0)$ only impacts the first term on the right hand side with $\mean_{x_0}[y(x_0)^2] = 1+ s^2$. Using further $q(t) =  \frac{\beta^T}{\sqrt d} \mean_{x_0} [x_0 z(x_0)^T] \frac{a_t}{\sqrt N}$ and $p(t) = \frac{a_t^T}{\sqrt N} \mean_{x_0} [z(x_0) z(x_0)^T] \frac{a_t}{\sqrt N}$, we write $\MSEtest_t = 1 + s^2 - 2 q(t) + p(t) $.

We provide analytical arguments to justify the formula \eqref{eq:test-error-decomposition} showing that:
\begin{gather}
    q(t) = \mu g(t) + o_d(1)\\
    p(t) = \mu^2 h(t) + \nu^2 l(t) + o_d(1)
\end{gather}
with 
\begin{align}
g(t) = \frac{\beta^T}{\sqrt d} \frac{\Theta^T}{\sqrt d}  \frac{a_t}{\sqrt N}, \quad
l(t) = \norm{\frac{a_t}{\sqrt N}}^2, \quad h(t) = \norm{\frac{\Theta^T}{\sqrt d} \frac{a_t}{\sqrt N}}^2
\end{align}
and where $\lim_{d\to +\infty} o_d(1) = 0$ with probability tending to one when $d\to +\infty$. The arguments below are based further on the prior assumption that the $(\theta_i/\sqrt{d})$ are sampled uniformly on the hyper-sphere of radius $1$. We will assume further that these results can be extended in our setting with $\theta_i$ sampled from a gaussian distribution. Notice that this is a reasonable assumption because $\norm{\theta_i}^2/d$ is a $\chi^2$ distribution of mean $1$ and variance $\frac{2}{d}$.

\subsection{limit of $q(t)$}
We decompose our activation function as $\sigma(x) = \mu x + \nu \sigma^\perp(x)$ where $\sigma^\perp \in \text{Span}(H_{e_i})_{i \geq 2}$. In other words, we have $\mean_G [\sigma^\perp(G)] = \mean_G [\sigma^\perp(G) G] = 0$ and $\mean_G [\sigma^\perp(G)^2] = 1$.  Notice that conditional on $(\theta_i)_i$ sampled on the sphere of radius $\sqrt{d}$, we have for all $i \in \{1, \ldots, N\}$ that $u_i \equiv \frac{\theta_i^Tx_0}{\sqrt d} \underset{x_0}{\sim} \mathcal N(0,1)$, and for all $j \in \{1, \ldots, N\}$, we have $\cov(u_i,u_j) = \frac{\theta_i^T \theta_j}{d} = \left[ \frac{\Theta \Theta^T}{d} \right]_{i,j}$. Similarly, for any $l \in \{1, \ldots, d\}$ we have $\cov(u_j, [x_0]_l) = \frac{[\theta_j]_l}{\sqrt d}$..
Now, using the Mehler-Kernel formula, we have
\begin{equation}
    \mean_{x_0}\left[
        [x_0]_l [z(x_0)]_{j}
    \right]
    =
    \sum_{k\geq 0}
    \frac{1}{k!}
    \left( 
        \cov(u_j, [x_0]_l)
    \right)^k
    \mean_{x_0} \left[ x_0 H_{e_k}(x_0) \right]
    \mean_{u_j} \left[ \sigma(u_j) H_{e_k}(u_j) \right]
\end{equation}
which does not vanish only for $k=1$ due to the first expectation on the RHS. Thus
\begin{equation}
    \mean_{x_0}\left[
        [x_0]_l [z(x_0)]_{j}
    \right]
    =  \frac{[\theta_j]_l}{\sqrt d} \mu
\end{equation}
and hence we find that $q(t) = \frac{\beta^T}{\sqrt d} \mean_{x_0} \left[ x_0 z(x_0)^T \right] \frac{a_t}{\sqrt N} = \mu \frac{\beta^T}{\sqrt d} \frac{\Theta^T}{\sqrt d} \frac{a_t}{\sqrt N}$. 

The result ought not be exact anymore when $(\theta_i)$ are sampled from a normal distribution, and we make the assumption that we can account for a correction term $o_d(1)$ which goes to $0$ as $d$ grows to infinity, hence $q(t) = \mu g(t) + o_d(1)$ in general.

\subsection{limit of $p(t)$}
Similarly for $p(t)$, we evaluate the kernel $U_{i,j} = \mean_{x_0}\left[
    [z(x_0)]_i [z(x_0)]_{j}
\right]$ for which the Mehler-Kernel formula provides
\begin{equation}
    \begin{array}{rcl}
       U_{i,j}
        & = &
        \sum_{k\geq 0}
        \frac{1}{k!}
        \left( 
            \cov(u_i, u_j)
        \right)^k
        \mean_{u_i} \left[ \sigma(u_i) H_{e_k}(u_i) \right]^2\\
        & = & \mu^2 \cov(u_i, u_j) + \nu^2 \sum_{k\geq 2}
        \frac{\left( 
            \cov(u_i, u_j)
        \right)^k}{k!}
        \mean_{u_i} \left[ \sigma^\perp(u_i) H_{e_k}(u_i) \right]^2 . \\
    \end{array}
\end{equation}
Intuitively, the terms $\left( \cov(u_i, u_j) \right)^k$ for $k \geq 2$ are on a smaller order in $d$ compared to $\cov(u_i,u_j)$ when $i \neq j$. We refer the reader to Lemma C.7 in \cite{mei2020generalization} where it is shown with some additional assumptions on $\sigma$ (weakly differentiable with $\exists c_0, c_1, \forall x>0, |\sigma(x)|, |\sigma'(x)| \leq c_0 e^{c_1 x}$) that:
\begin{equation}
    \mean_{\Theta} \left[
        \norm{
            U - \mu^2 \frac{\Theta \Theta^T}{d} - \nu^2 I_N
        }_{\text{op}}
    \right] = o_d(1) .
\end{equation}

Therefore, we can bound:
\begin{equation}
    \begin{array}[]{rcl}
        | p(t) - \mu^2 h(t) - \nu^2 l(t) | & = & \left\lvert
        \left\langle \frac{a_t^T}{\sqrt N} , \left(U - \mu^2 \frac{\Theta \Theta^T}{d}  - \nu^2\right) \frac{a_t^T}{\sqrt N} \right\rangle
        \right\rvert  \\
        & \leq & 
        \norm{ \frac{a_t}{\sqrt N}} \cdot
        \norm{
            U - \mu^2 \frac{\Theta \Theta^T}{d} - \nu^2 I_N
        }_{\text{op}} \cdot
        \norm{ \frac{a_t}{\sqrt N}} \\
        & = & l(t) \norm{
            U - \mu^2 \frac{\Theta \Theta^T}{d} - \nu^2 I_N
        }_{\text{op}} .
    \end{array}
\end{equation}
As per the general assumptions \ref{assumptions:highdim}, $l(t)$ concentrates to a finite quantity $\bar l(t)$ at all times as $d$ grows to infinity (that $\bar l(t)$ is finite is explicitly checked by the anlytical computations of the generalization error). Thus  by Markov's inequality we have at any fixed time $t$,  $| p(t) - \mu^2 h(t) - \nu^2 l(t) | = o_d(1)$ with probability tending to one as $d\to +\infty$.

Notice also that we assume as before that $o_d(1)$ also contains the correction added when $(\theta_i)$ are sampled from a normal distribution.

%% file: app-cauchy-integral-representation.tex
\section{Cauchy's integral representation formula}\label{app:cauchy-integral-repr}
In this section we complete the proof of propositions \ref{eq:cauchy_representation} and \ref{eq:cauchy_representation_train}. We show how to derive the Cauchy integral representation of the two functions $l(t)$ and $h(t)$ by similar analysis of Sect. \ref{sec:cauchyproof_g} for the representation of $g(t)$. 

\subsection{Representation formula for $l(t)$}\label{app:reprl}
We define the function $L_t(z) = \frac{a_t^T}{\sqrt N} R(z) \frac{a_t}{\sqrt N}$ and the auxiliary functions $U_t(z) = \frac{Y^T Z}{N} R(z) \frac{a_t}{\sqrt N}$ and $V(z) = \frac{Y^T Z}{N} R(z) \frac{Z^T Y}{N}$. We find a set of 2 integro-differential equations using the gradient flow equation for $\frac{\dd a_t}{\dd t}$ (as in the derivation of \ref{eq:d_Gt})
\begin{equation}\label{eq:pde_lt}
    \begin{array}{c}
        {\displaystyle \frac12 \frac{\partial L_t(z)}{\partial t} = U_t(z) - l(t) - (z+\delta) L_t(z)}\\[8pt]
        {\displaystyle \frac{\partial_t U_t(z)}{\partial t} = V(z) - \opR_z U_t - (z+\delta) U_t(z)}
    \end{array}
\end{equation}
Similarly $G_t(z)$ and $g(t)$, we also have that $l(t) = -\oint_\Gamma \frac{\dd z}{2 i \pi} L_t(z) = \opR_z L_t$. So we get a pair of integro-differential equations in this case (wheras for $G_t(z)$ we had only one such equation). However, we have one additional differential equation in this case. Pursuing with the Laplace transform operator\footnote{Defined as $(\opL f)(p) = \int_0^{+\infty} dt e^{-pt} f(t)$ for $\Re p$ large enough. We also use the notation $\opL f_p$ to mean 
$(\opL f)(p)$ specially when there are other variables involved. For example $\opL L_p(z) = \int_0^{+\infty} dt e^{-pt} L_t(z)$.}
the equations \eqref{eq:pde_lt} become
\begin{equation}
    \begin{array}{c}
        \opL L_p(z) = \frac{1}{\frac12 p + z + \delta} \left( \frac12 L_0(z) + \opL U_p(z) - \opL l(p) \right)\\[8pt]
        \opL U_p(z) = \frac{1}{p+z+\delta} \left( U_0(z) + \frac{V(z)}{p} - \opL \opR_z U_p  \right)
    \end{array}
\end{equation}
and re-injecting $\opL U_p$ from the second equation into the first equation we find
\begin{equation}
        \opL L_p(z) = \frac{1}{\frac12 p + z + \delta} \left( \frac{L_0(z)}{2} - \opL l(p) \right)
        + \frac{1}{(\frac12 p + z + \delta)(p+z+\delta)} \left( U_0(z) + \frac{V(z)}{p} - \opL \opR_z U_p  \right) .
\end{equation}
With similar considerations as before, with $p$ large enough to have $-\delta$ is outside the loop $\Gamma$, we see the terms $\opL l(p)$ and $ \opL \opR_z U_p$ don't contribute to the former equation when the operator $\opR_z$ is applied
\begin{equation}
    \opR_z \opL L_p(z) = \opR_z \left\{ \frac12 \frac{L_0(z)}{\frac12 p + z + \delta}
    + \frac{1}{(\frac12 p + z + \delta)(p+z+\delta)} \left( U_0(z) + \frac{V(z)}{p} \right)  \right\} .
\end{equation}
Finally, there remains to use the commutativity of $\opR_z$ and $\opL$ (for $\Re p$ large enough by Fubini's theorem) and compute the inverse Laplace transforms to find
\begin{equation}\label{eq:repr_l}
    l(t) = \opR_z \left\{ e^{-2t(z+\delta)} \left[ L_0(z)
    + 2 \frac{e^{t (\delta +z)}-1}{\delta +z} U_0(z) 
    + \left(\frac{e^{t (\delta +z)}-1}{\delta +z} \right)^2V(z)
    \right]
    \right\}
\end{equation}
Expanding further the terms individually
\begin{equation}
    l(t) = \opR_z \left\{ e^{-2t(z+\delta)} L_0(z)
    + 2 e^{-t(z+\delta)} \left( \frac{1 - e^{-t(z+\delta)} }{\delta +z} \right) U_0(z) 
    + \left(\frac{1-e^{-t (\delta +z)}}{\delta +z} \right)^2V(z)
    \right\} .
\end{equation}
We end-up (as for $g(t)$) with an expression where the time dependence is decoupled from random matrix expressions.

\subsection{Representation formula for $h(t)$}
The last term requires additional considerations. We will now use a double contour $\Gamma_x,\Gamma_y$ enclosing the eigenvalues of $\frac{Z^TZ}{\sqrt N}$ and such that $\Gamma_x \cap \Gamma_y = \emptyset $. We consider the operators $\opR_x,\opR_y$ associated to each contour.
Contrary to the previous two representations, when computing the multiple derivatives $h^{(k)}(t)$, 
due to the $\Theta$ matrix in $h(t)$, there appears pairs of matrices $\frac{Z^TZ}{\sqrt N}$. 
In terms of generating functions, this translates into a "2-variable resolvent" functions 
\begin{align}
H_t(x,y) = \frac{a_t^T}{\sqrt N} R(x) \frac{\Theta \Theta^T}{d} R(y) \frac{a_t}{\sqrt N} ,
\end{align}
which has the property $h(t) = \opR_{x,y} H_t$, and two auxiliary functions 
\begin{align}
Q_t(x,y) = \frac{a_t^T}{\sqrt N} R(x) \frac{\Theta \Theta^T}{d} R(y) \frac{Z^T Y}{N} , 
\quad 
{\rm and} 
\quad 
W(x,y) = \frac{Y^T Z}{N} R(x) \frac{\Theta \Theta^T}{d} R(y) \frac{Z^T Y}{N} .
\end{align}
Using the former method for equation \eqref{eq:pde_lt} leads to the following integro-differential equations:
\begin{equation}\label{eq:h_t_diff}
    \begin{array}{l}
        {\displaystyle
        \frac{\partial H_t(x,y)}{\partial t}
        = Q_t(x,y) + Q_t(y,x) - \opR_x H_t(y) - \opR_y H_t(x) - (x+y+2\delta) H_t(x,y)} \\[8pt]
        {\displaystyle \frac{\partial Q_t(x,y)}{\partial t}
        = W(x,y) - \opR_x Q_t(y) - (x + \delta) Q_t(x,y)}
    \end{array}
\end{equation}
Then the Laplace transform on the first equation reads
\begin{equation}\label{eq:H_t}
    \opL H_p(x,y) 
    = \frac{1}{p + x+y+2\delta} \left[ H_0(x,y) + \opL \left\{ 
        Q_t(x,y) + Q_t(y,x) - \opR_x H_t(y) - \opR_y H_t(x)
    \right\} \right] .
\end{equation}
Notice that $\opR_x$ and $\opR_y$ commute with each other as being integrals over a compact set $\Gamma_x,\Gamma_y$ respectively. So by Fubini we can name indifferently $\opR_{x,y} = \opR_x \opR_y = \opR_y \opR_x$. Notice also that $\opR_x H_t(y)$ is not a function of $x$ anymore, thus for $p$ large enough to have $|2\delta+x+y|>0$ for all $(x,y) \in \Gamma_x \times \Gamma_y$, we find
\begin{equation}\label{eq:0simp}
    \opR_{x,y} \left\{ \frac{\opR_x H_t(y)}{p + x+y+2\delta} \right\}
    = \opR_y \left\{ \opR_x \left\{ \frac{\opR_x H_t(y)}{p + x+y+2\delta} \right\} \right\}
    = \opR_y \{0\} = 0 .
\end{equation}
Symmetrically, the same statement can be made for $\opR_y H_t(x)$, so applying the operator $\opR_{x,y}$ and the result \eqref{eq:0simp} to \eqref{eq:H_t} we find 
\begin{equation}\label{eq:H_t_2}
    \opR_{x,y} \opL H_p(x,y) 
    = \opR_{x,y} \left\{ 
        \frac{H_0(x,y) + \opL Q_p(x,y) + \opL Q_p(y,x)}{p + x+y+2\delta} 
    \right\} .
\end{equation}
Finally, we have $\opR_{x,y} \opL H_p(x,y) = \opL \opR_{x,y}  H_p(x,y) = \opL h(p)$. The Laplace transform of the second equation of \eqref{eq:h_t_diff} provides 
\begin{equation}
    \opL Q_p(x,y) = \frac{1}{p+x+\delta} \left( Q_0(x,y) + \frac{W(x,y)}{p} - \opR_x \opL Q_p(y) \right) .
\end{equation}
Before injecting this equation into \eqref{eq:H_t_2} (and its symmetrical result in $x$ and $y$), notice that one term will not contribute under the operator $\opR_{x,y}$
\begin{equation}
    \opR_{x,y} 
    \left\{ \frac{\opR_x \opL Q_p(y)}{(p + x+y+2\delta)(p+x+\delta)} \right\}
    = \opR_y \{0\} = 0
\end{equation}
and finally, using  $W(x,y) = W(y,x)$, we obtain
\begin{equation}
    \opL h(p)
    = \opR_{x,y} \left\{ 
        \frac{1}{p + x+y+2\delta} \left( H_0(x,y) +
        \frac{Q_0(x,y)+ \frac{W(x,y)}{p} }{p+x+\delta} 
        + \frac{Q_0(y,x)+ \frac{W(x,y)}{p} }{p+y+\delta} 
        \right)
    \right\} .
\end{equation}
Eventually, applying inverse Laplace transform we get the representation
\begin{equation}
    \begin{array}{rcl}
        h(t) & = & 
        \opR_{x,y} \left\{ e^{-t(x+y+2\delta)} H_0(x,y) 
        \right\} \\
        & + & \opR_{x,y} \left\{ 
            e^{-t (2 \delta +x+y)} \left(
            \frac{e^{ t (\delta + y) }- 1}{\delta + y} Q_0(x,y) +
            \frac{e^{ t (\delta + x) }- 1}{\delta + x} Q_0(y,x)
            \right) \right\}\\
            & + & \opR_{x,y} \left\{ 
                \frac{1-e^{-t(x+\delta)}}{x+\delta}
                \frac{1-e^{-t(y+\delta)}}{y+\delta}
             W(x,y) \right\}
    \end{array}
\end{equation}

\subsection{Remark on the consistency with the minimum least squares estimator} 
It can be seen, at least formally, that the integral representation formula correctly retrieves the minimum least-squares estimator formulas in the limit $t \to \infty$. Indeed, commuting $\lim_t$ and $\opR_z$ we find 
\begin{align}
\lim_{t\to +\infty} g(t) & = \opR_z \left\{ \frac{1}{z+\delta} K(z) \right\} = \sum_{i=1}^N \frac{\beta^T}{\sqrt d} v_i \opR_z \left\{ \frac{1}{(\lambda_i+z)(z+\delta)} \right\} v_i^T \frac{Z^T Y}{N} 
\nonumber \\ &
= \sum_{i=1}^N \frac{\beta^T}{\sqrt d} \frac{v_i v_i^T}{(\lambda_i-\delta)}  \frac{Z^T Y}{N} = K(-\delta) .
\end{align}
On the other hand, we expect 
\begin{align}
\lim_{t\to +\infty} g(t) = \lim_t \frac{\beta^T}{\sqrt d} \frac{\Theta^T}{\sqrt d} a_t = \frac{\beta^T}{\sqrt d} \frac{\Theta^T}{\sqrt d} a_\infty
\end{align} 
with $a_\infty$ defined as the minimum least-squares estimator. Thus, we clearly have: 
\begin{equation}
    \frac{\beta^T}{\sqrt d} \frac{\Theta^T}{\sqrt d} a_\infty = \frac{\beta^T}{\sqrt d} \frac{\Theta^T}{\sqrt d} \left(\frac{Z^TZ}{N} + \delta I\right)^{-1} \frac{Z^T}{\sqrt N} \frac{Y}{\sqrt N} = K(-\delta)
\end{equation}
The same calculations can be done on each term $h(t),l(t)$.

\subsection{Representation formula for the training error}
The derivation of $\MSEtrain_t$ is quite straightforward based on the previous terms derived for the test error. Firstly, expanding the expression of $\MSEtrain_t$ we get:
\begin{equation}
    \MSEtrain_t = \frac{1}{n} \norm{ Y - Z \frac{a_t}{\sqrt N} }^2 + \lambda \norm{\frac{a_t}{\sqrt N}}^2 = 
    \frac{\norm{Y}^2}{n}
    - \frac{2}{n} Y^T \frac{Z a_t}{\sqrt N}
    + \frac{1}{n} \norm{\frac{Z a_t}{\sqrt N}}^2
    + \frac{\delta}{c} \norm{\frac{a_t}{\sqrt N}}^2
\end{equation}

Reusing the function $U_t(z)$ from Sect. \ref{app:reprl}, and defining $u(t) = \opR_z U_t(z) = \frac{1}{N}  Y^T \frac{Z a_t}{\sqrt N} $ and $\tilde h(t) = \frac{1}{N} \norm{ \frac{Z a_t}{\sqrt N}}^2$, we get:
\begin{equation}\label{eq:msetrain}
    \MSEtrain_t = \frac{\norm{Y}^2}{n} + \frac{1}{c} \left( - 2 u(t) + \tilde h(t) + \delta l(t) \right)
\end{equation}
Furthermore, reusing the differential equation found for $U_t(z)$, a simpler solution can be extracted for $u(t)$:
\begin{equation}
    u(t) = \opR_z \left\{
        e^{-t(z+\delta)} U_0(z) + \frac{1-e^{-t(z+\delta)}}{z + \delta} V(z)
     \right\}
\end{equation}
The second term $\tilde h(t)$ can also be derived from the expression $L_t(z)$ which is also defined in appendix \ref{app:reprl}. We find $\tilde h(t) = \opR_z \{ z L_t(z) \}$. Hence the terms $\delta l(t)$ and $\tilde h(t)$ can be grouped together with $\tilde h(t) + \delta l(t) = \opR_z \{ (z+\delta) L_t(z) \}$. Expanding from the expression of $\opR_z \opL L_t(z)$ we find
\begin{align}
    (\tilde h + \delta l)(t) = 
    \opR_z 
    \biggl\{ & 
    (z+\delta)e^{-2t(z+\delta)} L_0(z)
     + 2 e^{-t(z+\delta)} \left( 1 - e^{-t(z+\delta)} \right) U_0(z) 
    \nonumber \\ &
    + \frac{\left( 1-e^{-t (\delta +z)} \right)^2}{\delta +z} V(z)
    \biggr\} .
\end{align}
Remarkably, all the terms can be summed together in \eqref{eq:msetrain} and we retrieve a simpler expression
\begin{equation}
        \MSEtrain_t \! = \! \frac{\norm{Y}^2}{n} + \frac{1}{c} \opR_z 
        \left\{ \! (z+\delta) e^{-2t(z+\delta)} L_0(z)
        - 2 e^{-2t(z+\delta)} U_0(z) - \frac{1 - e^{-2 t (\delta +z)}}{\delta +z} V(z)
        \!
        \right\} .
\end{equation}

%% file: app-random-matrix-limits.tex
\section{High-dimensional limit}

In this appendix we use assumption \ref{assumptions:highdim} in section \ref{subsec:highdim} to compute limiting expressions of traces.

As $d\to\infty$, the mean of $a_0$ or $\beta$ converges two $0$. Let's consider the auxiliary functions $U_0(z), G_0(z), Q_0(x,y)$. These three terms have only occurrence of $a_0$ and $\beta$ on each side of the matrix-vector multiplication composition (notice $\beta$ is also included in the term $Y$): they can be written in the form $F(H) = \frac{a_0^T}{\sqrt N} H \frac{\beta}{\sqrt d}$ where $H$ is a random matrix independent of $a_0, \beta$. For instance we have $G_0(z) = F\left(R(z) \frac{\Theta}{\sqrt d} \right)$. As the mean of $F(H)$ is precisely $0$, assuming concentration, we have that these terms go to $0$ when $d \to \infty$. The same considerations can be applied to the term $\xi$ from $Y$.

Besides, when a vector such as $a_0$ is expressed on both side of another expression such as $F(H) = \frac{a_0^T}{\sqrt N} H \frac{a_0}{\sqrt N}$, it can still be rewritten as the trace $F(H) = \trace{H \frac{a_0 a_0^T}{N}}$ so that we can effectively use the independence of $H$ with $a_0$ and compute the expectation $\mean_{a_0} [F(H)] = \frac{r^2}{N} \trace{H}$. Hence if $F(H)$ concentrates as $N \to \infty$, we can replace it by $\lim_N \frac{r^2}{N} \trace{H}$.

In the sequel we will adopt the following notation. For any sequence of matrices $(M_k) \in \mathbb{R}^{k\times k}$ we set $\traceLim[k]{M_k} = \lim_{k \to \infty} \frac{1}{k} \trace{M_k}$. 

Therefore, in general, applying the concentration arguments above, we can substitute the limiting expressions with the following terms
\begin{gather}\label{eq:list_eq}
    L_0(z) = \frac{a_0^T}{\sqrt N} R(z) \frac{a_0}{\sqrt N} 
    \underset{d \to \infty}{\longrightarrow}
    r^2 ~\traceLim[N]{R(z)}\\
    K(z) = \frac{\beta^T}{\sqrt d} \frac{\Theta^T}{\sqrt d} R(z) \frac{Z^T Y}{N}
    \underset{d \to \infty}{\longrightarrow} 
     \traceLim[d]{ \frac{\Theta^T}{\sqrt d} R(z) \frac{Z^T}{\sqrt N} \frac{X}{\sqrt N} } \\
    H_0(x,y) = \frac{a_0^T}{\sqrt N} R(x) \frac{\Theta \Theta^T}{d} R(y) \frac{a_0}{\sqrt N}
    \underset{d \to \infty}{\longrightarrow}
    r^2 ~ \traceLim[N]{ R(x) \frac{\Theta \Theta^T}{d} R(y) }
    \\
    V(z) = \frac{Y^T Z}{N} R(z) \frac{Z^TY}{N}
    \underset{d \to \infty}{\longrightarrow}
    \traceLim[d]{ \frac{X^T}{\sqrt N} \frac{Z}{\sqrt N} R(z) \frac{Z^T}{\sqrt N} \frac{X}{\sqrt N}} + s^2 \traceLim[N]{ \frac{Z}{\sqrt N} R(z) \frac{Z^T}{\sqrt N} }
    \\
    W(x,y) 
    \underset{d \to \infty}{\longrightarrow}
    \traceLim[d]{ \frac{X^T}{\sqrt N} \frac{Z}{\sqrt N} R(x) \frac{\Theta \Theta^T}{d} R(y) \frac{Z^T}{\sqrt N} \frac{X}{\sqrt N} }
    + s^2 \traceLim[N]{ \frac{Z}{\sqrt N} R(x) \frac{\Theta \Theta^T}{d} R(y) \frac{Z^T}{\sqrt N} }
\end{gather}

As for the training error, all the required terms are given by $V(z), L_0(z), U_0(z)$, of which
only $V(z), L_0(z)$ contributes to the result as $d\to \infty$

Finally, we apply the gaussian equivalence principle with the substitution described in \ref{def:gausseq} with the linearization $Z \to \Zlin$ with $\Zlin \equiv \frac{\mu}{\sqrt d} X \Theta^T + \nu \Omega$. This substitution is applied throughout all the occurrences of $Z$, including in the resolvents $z \to R(z)$.

%% file: app-linear-pencil.tex
\section{Linear Pencil}\label{app:linear-pencil}

\subsection{Main matrix}
The main approach of the linear-pencil method is to design a block-matrix $M_{x,y} = \sum_{i,j} E_{i,j} \otimes M_{x,y}^{(i,j)}$ where the blocks $M_{x,y}^{(i,j)}$ are either a gaussian random matrix or a scalar matrix, and $E_{i,j}$ is the matrix with matrix elements $(E_{i,j})_{k,l} = \delta_{ki}\delta_{lj}$. The subscripts indicate explicitly the dependence on two complex variables $(x, y) \in \mathbb{C}^2$. Importantly, this matrix is inverted using block-inversion formula to have an expression of the form $M_{x,y}^{-1} = \sum_{i,j} E_{i,j} \otimes (M_{x,y}^{-1})^{(i,j)}$ such that some blocks $(M_{x,y}^{-1})^{(i,j)}$ match the different matrix terms in equations \eqref{eq:list_eq}.

In order to define our main linear pencil matrix, we first need to introduce some additional upper-level blocks:
$U^T = [ \frac{X}{\sqrt N}, \nu \frac{\Omega}{\sqrt N}]$ and $V^T = [\mu \frac{\Theta}{\sqrt d}, I]$. In addition, in order to keep a consistent symmetry and structure to our block-matrix, we will use the following blocks in reverse order: $\bar U^T = [\nu \frac{\Omega}{\sqrt N}, \frac{X}{\sqrt N}]$ and $\bar V^T = [I, \mu \frac{\Theta}{\sqrt d}]$. Furthermore, we let $K_x = (-xI + \frac{\Zlin^T\Zlin}{\sqrt N})^{-1}$ and $L_x = (-xI + UU^TVV^T)^{-1}$ and $R_x = (-xI + VV^TUU^T)^{-1}$ and $\tilde K_x = (-xI + \frac{\Zlin\Zlin^T}{\sqrt N})^{-1}$.
The following identities (which can be obtained with the push-through identity) provide additional relations which can be used later:
\begin{eqnarray}
    \frac{\Zlin}{\sqrt N} = U^TV\\
    L_x U U^T = U \tilde K_x U^T\\
    V V^T L_x = V K_x V^T\\
    -x \tilde K_x = I - \left(-xI + \frac{\Zlin\Zlin^T}{ N}\right)^{-1} \frac{\Zlin\Zlin^T}{N}
    = I - \frac{\Zlin}{\sqrt N} K_x \frac{\Zlin^T}{\sqrt N} 
\end{eqnarray}

We define our main block-matrix consisting in $13\times 13$ blocks where the upper-level blocks $U,V,\bar U, \bar V$ are to be considered as "flattened":
\begin{equation}\label{eq:real_mxy}
    M_{x,y} =\left[
        \begin{array}{c|ccc||c|ccc|c}
            -xI & -V^T & 0 & 0 & \frac{\Theta}{\sqrt d}  & 0 & 0 & 0 & 0\\ \hline
            0 & I & U & 0 & 0 & 0 & 0 & 0 & 0\\
            0 & 0 & I & U^T & 0& 0 & 0 & 0 & 0\\
            V & 0 & 0 & I & 0 & 0 & 0 & 0 & 0\\ \hline \hline
            0 & 0 & 0 & 0 & I & 0 & 0 & 0 & \frac{\Theta^T}{\sqrt d} \\ \hline 
            0 & 0 & 0 & 0 & 0 & I & \bar U & 0 & 0\\ 
            0 & 0 & 0 & 0 & 0 & 0 & I & \bar U^T & 0\\ 
            0 & 0 & 0 & 0 & 0 & 0 & 0 & I & -\bar V\\ \hline
            0 & 0 & 0 & 0 & 0 & \bar V^T & 0 & 0 & -y I\\ 
    \end{array} \right]
\end{equation}
This is precisely the block-matrix $M$ given at the end of Sect. \ref{sec:sketch-proof}.

\subsection{Linear-pencil inversion and relation to the matrix terms}
The inverse of $M_{x,y}$ can be computed by splitting it into higher-level blocks. These blocks are highlighted with the lines and double-lines depicted in equation \eqref{eq:real_mxy}: the block-matrix is split into a $2 \times 2$ block-matrix recursively in order to apply the block-matrix inversion formula recursively. Starting with the higher level split:
\begin{equation}
    M_{x,y} =\left[
        \begin{array}{c||c}
            M_{1,1} & M_{1,2} \\ \hline \hline
            0 & M_{2,2}
    \end{array} \right]
    \Longrightarrow
    M_{x,y}^{-1} =\left[
        \begin{array}{c||c}
            M_{1,1}^{-1} & - M_{1,1}^{-1} M_{1,2} M_{2,2}^{-1}\\ \hline \hline
            0 & M_{2,2}^{-1}
    \end{array} \right]
\end{equation}
It is now quite straightforward algebra to proceed with the remaining blocks. Starting with $M_{1,1}$:
\begin{equation}
    M_{1,1}^{-1} =\left[
        \begin{array}{cccc}
            K_x & K_x V^T & -K_x \frac{\Zlin^T}{\sqrt N} & K_x \frac{\Zlin^T}{\sqrt N} U^T\\ 
            - U \frac{\Zlin}{\sqrt N} K_x & -xL_x & x L_x U & -x L_x U U^T\\
            \frac{\Zlin}{\sqrt N} K_x & \frac{\Zlin}{\sqrt N} V^T L_x & -x \tilde K_x & x \tilde K_x U^T\\
            - V K_x &  -V V^T L_x & V \frac{\Zlin^T}{\sqrt N} \tilde K_x & -x R_x 
    \end{array} \right]
\end{equation}
For $M_{2,2}$, with an additional split:
\begin{equation}
    M_{2,2} = \left[
        \begin{array}{c|c}
            I & N_{1,2}\\ \hline
            0 & N_{2,2}
    \end{array} \right]
    \Longrightarrow
    M_{2,2}^{-1} = \left[
        \begin{array}{c|c}
            I & -N_{1,2} N_{2,2}^{-1}\\ \hline
            0 & N_{2,2}^{-1}
    \end{array} \right]
\end{equation}
A straightforward algebra calculation provides the result of $M_{2,2}^{-1}$:
\begin{equation}
    M_{2,2}^{-1} = \left[
        \begin{array}{ccccc} 
            I & \frac{\Theta^T}{\sqrt d} K_y \bar V^T & -\frac{\Theta^T}{\sqrt d} K_y \frac{\Zlin^T}{\sqrt N}  & \frac{\Theta^T}{\sqrt d} K_y \frac{\Zlin^T}{\sqrt N} \bar U^T & -\frac{\Theta^T}{\sqrt d} K_y \\
            0 & -y \bar R_y & y \bar U \tilde K_y & -y \bar U \bar U^T \bar L_y & \bar U \frac{\Zlin}{\sqrt N} K_y \\
            0 & \tilde K_y \frac{\Zlin}{\sqrt N} \bar V^T  & -y \bar K_y & y \bar U^T \bar L_y & -\frac{\Zlin}{\sqrt N} K_y \\
            0 & - \bar L_y \bar V \bar V^T & \bar L_y \bar V \frac{\Zlin^T}{\sqrt N} & -y \bar L_y & \bar V K_y \\
            0 & -K_y \bar V^T & K_y \frac{\Zlin^T}{\sqrt N}  & -K_y \frac{\Zlin^T}{\sqrt N} \bar U^T & K_y \\
    \end{array} \right]
\end{equation}
Finally, using $Q = K_x \frac{\Theta \Theta^T}{d} K_y$ we obtain the third block of $M_{x,y}$:
\begin{equation}
    \begin{array}{c}
    
    -M_{1,1}^{-1} M_{1,2} M_{2,2}^{-1} =
    \left[
        \begin{array}{ccccc} 
            - K_x \frac{\Theta}{\sqrt d} & -Q \bar V^T & Q \frac{\Zlin^T}{\sqrt N} & -Q \frac{\Zlin^T \bar U^T}{\sqrt N} & Q\\ 
            \frac{U\Zlin}{\sqrt N} K_x \frac{\Theta}{\sqrt d} & \frac{U\Zlin}{\sqrt N} Q \bar V^T & - \frac{U\Zlin}{\sqrt N} Q \frac{\Zlin^T}{\sqrt N} &  \frac{U\Zlin}{\sqrt N} Q \frac{\Zlin^T \bar U^T}{\sqrt N} & - \frac{U\Zlin}{\sqrt N} Q\\
            - \frac{\Zlin}{\sqrt N} K_x \frac{\Theta}{\sqrt d} & - \frac{\Zlin}{\sqrt N} Q \bar V^T &  \frac{\Zlin}{\sqrt N} Q \frac{\Zlin^T}{\sqrt N} & -\frac{\Zlin}{\sqrt N} Q \frac{\Zlin^T \bar U^T}{\sqrt N}& \frac{\Zlin}{\sqrt N} Q\\
            V K_x \frac{\Theta}{\sqrt d} & V Q \bar V^T & - V Q \frac{\Zlin^T}{\sqrt N} & V Q \frac{\Zlin^T \bar U^T}{\sqrt N} & -V Q
        \end{array} 
        \right]
    \end{array}
\end{equation}
Notice now that all the matrix terms in equations \eqref{eq:list_eq} are actually contained in some of the blocks of our matrix (note that $\traceLim[d]{ \frac{X^TX}{n}}=1$):
\begin{gather}\label{eq:structures2}
    \bar L_0(y) = r^2 \traceLim[N]{K_y}\\
    \bar K(y) =  \traceLim[d]{ \frac{\Theta^T}{\sqrt d} K_y \frac{\Zlin^T}{\sqrt N} \bar U^T }_{1,2}\\
    \bar H_0(x,y) = r^2 \traceLim[N]{Q}\\
    \bar W(x,y) = s^2 \frac{\phi}{\psi} \traceLim[n]{\frac{\Zlin}{\sqrt N} Q \frac{\Zlin^T}{\sqrt N}} + \traceLim[d]{\frac{U\Zlin}{\sqrt N} Q \frac{\Zlin^T \bar U^T}{\sqrt N}}_{1,2}\\
    \bar V(x) = 
    s^2 \frac{\phi}{\psi} \traceLim[n]{I_n+x \tilde K_x} +
    \left( 
        \traceLim[d]{x L_x U U^T}_{1,1}
        +
        \traceLim[d]{ \frac{X^TX}{N}}
    \right)
\end{gather}
Or equivalently, with the block coordinates of the inverse matrix $M_{x,y}^{-1}$:
\begin{gather}\label{eq:structures3}
    \bar L_0(y) = r^2 \traceLim[N]{ (M_{x,y}^{-1})^{(13,13)} }\\
    \bar K(y) =  \traceLim[d]{ (M_{x,y}^{-1})^{(7,12)} }\\
    \bar H_0(x,y) = r^2 \traceLim[N]{ (M_{x,y}^{-1})^{(1,13)} }\\
    \bar W(x,y) = s^2 \frac{\phi}{\psi} \traceLim[n]{ (M_{x,y}^{-1})^{(4,10)} } + \traceLim[d]{ (M_{x,y}^{-1})^{(2,12)} }\\
    \bar V(x) = 
    s^2 \frac{\phi}{\psi} \left(1 - \traceLim[n]{ (M_{x,y}^{-1})^{(4,4)} }\right) +
    \left( 
        -\traceLim[d]{ (M_{x,y}^{-1})^{(2,5)} }
        +
        \frac{\phi}{\psi}
    \right)
\end{gather}
In the next section we show how to derive further each trace of the squared matrices from the block matrix $M_{x,y}$. In order to deal with self-adjoint matrices, we double the dimensions with $\tilde M_{x,y}$:
\begin{equation}
    \tilde M_{x,y} = \left[
    \begin{array}{cc}
        0 & M_{x,y}\\
        M_{x,y}^\dag & 0
    \end{array} \right]
\end{equation}
and find the inverse:
\begin{equation}
    \tilde M_{x,y}^{-1} = \left[
    \begin{array}{cc}
        0 & (M_{x,y}^\dag)^{-1}\\
        M_{x,y}^{-1} & 0
    \end{array} \right]
\end{equation}

\subsection{Structural terms of the limiting traces}
% Est-ce qu'il faudrait donner un autre nom que eta pour ne pas confondre avec le learning rate ? je ne crois pas qu'on peut confondre en fait (N).
%As required in the theorem from \cite{mingo2017free}, 
The matrix $M_{x,y}$ is a block-matrix constituted with either gaussian random matrices, or constant matrices (proportional to $I$). More precisely, letting $S$ be the matrix of the coefficients of the constant blocks of $M_{x,y}$ (and $\tilde S$ for $\tilde M_{x,y}$), and $A$ the random blocks part ($\tilde A$ respectively) we write : $\tilde M_{x,y} = \sum_{i,j} E_{i,j} \otimes \tilde M_{x,y}^{(i,j)}$ where $\tilde M_{x,y}^{(i,j)} = \tilde S^{(i,j)} + \tilde A^{(i,j)}$ is the block of size $(N_i,N_j)$. Also notice that letting $\mathbb L = \{ (i,j) | N_i = N_j\}$, the fact that the constant blocks are supposed to be proportional to an identity matrix implies that: $\forall (i,j) \notin \mathbb L \implies \tilde S^{(i,j)} = 0 = z_{i,j} 0_{N_i,N_j}$  with $0_{N_i,N_j}$ the zero-matrix of size $N_i\times N_j$ and otherwise $\forall (i,j) \in \mathbb L \implies \tilde S^{(i,j)} = z_{i,j} I_{N_i}$ with $\tilde B = (z_{i,j})$ the matrix of size $26\times 26$.

Now we want to find a matrix $\tilde G \in \mathbb{R}^{26\times 26}$ such that  
\begin{align}\label{equ:requirement}
[\tilde G]_{i,j} = \traceLim[N_i]{ (\tilde M_{x,y}^{-1})^{(i,j)} }, \quad \forall (i,j) \in \mathbb L,
\end{align}
An important theorem in \cite{mingo2017free} (chapter 9, equ. (9.5) and theorem 2), which we show again in the next section, states that there is a solution $\tilde G$ of the equation 
\begin{equation} \label{eq:general_eq}
    \tilde B \tilde G = I + \eta(\tilde G) \tilde G
\end{equation}
which satisfies \eqref{equ:requirement}. In this equation 
$\eta(\tilde G)$ is the matrix mapping defined element-wise as: 
\begin{gather}
    [\eta(\tilde G)]_{i,j} = \delta_{\mathbb L} (i,j) \cdot \sum_{k,l \in \mathbb L} \sigma(i,k;l,j) \cdot [\tilde G]_{k,l}
\end{gather}
and where $\sigma$ satisfies the relation for all $(i,k,l,j)$ such that $N_i = N_j$ and $N_k = N_l$ (and keeping in mind that the $N_k$ are growing with the dimension $d$):
\begin{equation}\label{equ:sigmadef}
    \forall (r,s) \in \{ 1, \ldots, N_i\}\times \{1, \ldots, N_j\},
    r \neq s \implies 
    \sigma(i,k;l,j) = \lim_{d \to \infty} N_k \cdot \mean \left[  [\tilde A^{(i,k)}]_{r,s} [\tilde A^{(l,j)}]_{s,r}  \right]
\end{equation}
We remark that the setting here, and in particular equation \eqref{eq:general_eq}, is in fact more general than in \cite{mingo2017free} (chapter 9, equ. (9.5)) and we provide an independent and self-contained (formal) derivation of \eqref{eq:general_eq} in Appendix \ref{app:replica} using the replica method.

For example, we have $M_{x,y}^{(5,1)} = \mu \frac{\Theta^T}{\sqrt d}$ of size $d\times{N}$ and $M_{x,y}^{(1,7)} = \frac{\Theta}{\sqrt d}$ of size $N\times{d}$. So this is $\tilde M_{x,y}^{(5,14)} = \mu \frac{\Theta^T}{\sqrt d}$ and $\tilde M_{x,y}^{(1,20)} = \frac{\Theta}{\sqrt d}$, with $N_5 = N_{20} = d$ and $N_{14} = N_{1} = N$. For $r=1,s=2$ (or any other suitable indices) we find: 
$$
\sigma(5,14;1,20) = 
\lim_{d \to \infty} \mu \frac{N}{d} \mean \left[ [\Theta]_{1,2}^2 \right]
= \mu \psi
$$
In fact, a careful inspection of all the blocks in row $5$ and all the blocks in column $20$ shows that we have $[\eta(\tilde G)]_{5,20} = \mu \psi [\tilde G]_{14,1}$. 

Calculating all the terms of $\eta(\tilde G)$ is quite cumbersome, but it can be done automatically with the help of a computer algebra system. Still, this approach yields many equations for each $26\times{26}$ terms of $\tilde G$. However, some initial structure can also be provided for this matrix. Looking back at $\tilde M_{x,y}^{-1}$, it is clear that some blocks will have the same limiting traces (potentially seen using the aforementioned push-through identities). For instance, $(M_{1,1}^{-1})^{(1,1)} = K_x = -(M_{1,1}^{-1})^{(6,1)}$ (expanding the $U,V$ blocks), so $(M_{x,y}^{-1})^{(1,1)} = -(M_{x,y}^{-1})^{(6,1)}$, in other words $(\tilde M_{x,y}^{-1})^{(14,1)} = -(\tilde M_{x,y}^{-1})^{(19,1)}$, and thus we expect $[\tilde G]_{14,1} = -[\tilde G]_{19,1}$.
Non-squared blocks can also be mapped to $0$ in $\tilde G$. In the end, taking every block into account, $\tilde G$ is expected to be of the form:
\begin{equation}
    \tilde G = \left[
        \begin{array}{c|c}
            0 & G^\dag\\ \hline
            G & 0
        \end{array} \right]
\end{equation}
with 
\begin{equation}
    G = \left[
    \begin{array}{c|c|c}
        G_{1,1} & G_{1,2} & G_{1,3}\\ \hline
        0 & 1 & G_{2,3}\\ \hline
        0 & 0 & G_{3,3}
    \end{array} \right]
\end{equation}
(which has $13\times 13$ scalar matrix elements)
where:
\begin{equation}
    G_{1,3} =
    \left[
    \begin{array}{cc|c|cc|c}
        -q_1 & 0 & 0 & -\nu q_6^{yx} & 0 & q_1\\ \hline
        0 & \mu q_7^{yx} & 0 & 0 & q_2 & 0\\ 
        \nu q_6^{xy} & 0 & 0 & \nu^2 q_3 & 0 & -\nu q_6^{xy} \\ \hline
        0 & 0 & q_4 & 0 & 0 & 0\\ \hline
        0 & \mu^2 q_5 & 0 & 0 & \mu q_7^{xy}  & 0\\ 
        q_1 & 0 & 0 & \nu q_6^{yx} & 0 & -q_1\\ 
    \end{array} 
    \right]
\end{equation}
\begin{equation}
    G_{1,1} =
    \left[
    \begin{array}{c|cc|c|cc}
        g_1^x & 0 & g_1^x & 0 & 0 & \nu g_2^x  \\ \hline
        0 & h_1^x & 0 & 0 & h_4^x & 0\\
        -\nu g_2^x & 0 & h_2^x & 0 & 0 & \nu^2 h_5^x \\ \hline
        0 & 0 & 0 & g_3^x & 0 & 0\\ \hline
        0 & -\mu^2 h_3^x & 0 & 0 & h_1^x & 0\\ 
        -g_1^x & 0 & -g_1^x & 0 & 0 & h_2^x
    \end{array} 
    \right]
\end{equation}
\begin{equation}
    G_{3,3} =
    \left[
    \begin{array}{cc|c|cc|c}
        h_2^y & 0 & 0 & \nu^2 h_5^y & 0 & \nu g_2^y\\
        0 & h_1^y & 0 & 0 & h_4^y & 0\\ \hline
        0 & 0 & g_3^y & 0 & 0 & 0\\ \hline
        -g_1^y & 0 & 0 & h_2^y & 0 & g_1^y\\
        0 & -\mu^2 h_3^y & 0 & 0 & h_1^y & 0\\ \hline
        -g_1^y & 0 & 0 & -\nu g_2^y & 0 & g_1^y\\
    \end{array} 
    \right]
\end{equation}
\begin{equation}
    G_{1,2} =
    \left[
    \begin{array}{c}
        0 \\ \hline
        t_1^x \\
        0 \\ \hline
        0 \\ \hline
        \mu h_3^x \\ 
        0 \\
    \end{array} 
    \right] ~~~~
    G_{2,3} =
    \left[
    \begin{array}{cc|c|cc|c}
        0 & \mu h_3^y & 0 & 0 & t_1^y & 0 
    \end{array} 
    \right]
\end{equation}
All (non-vanishing) matrix elements depend on the complex variables $x$ and $y$. This is indicated by 
the upper-script notation with $x,y,xy,yx$. Some quantities depend only on $x$, some only on $y$, and some on both $x$ and $y$. Among the ones that depend on both variables the quantities $q_6^{xy},q_6^{yx},q_7^{xy},q_7^{yx}$ are {\it non-symmetric}, while $q_1,q_2,q_3,q_4,q_5$ are {\it symmetric} (e.g.,  $q_1^{x,y} = q_1^{y,x}$). We choose not to use the upper-script notation for the symmetric quantities in order to distinguish them from the {\it non-symmetric} ones.

Eventually, with a careful mapping between $\tilde M_{x,y}^{-1}$ and $\tilde G$ in equations \eqref{eq:structures2}, only $g_1^x, t_1^x, h_4^x, g_3^x$ and the symmetric terms $q_1, q_2, q_4$ are needed and equations \eqref{eq:structures2} take the form:
\begin{gather}
    \bar L_0(x) = r^2  g_1^x\\
    \bar K(x) = t_1^x\\
    \bar H_0(x,y) = r^2 q_1\\
    \bar W(x,y) = s^2 \frac{\phi}{\psi} q_4 + q_2\\
    \bar V(x) = s^2 \frac{\phi}{\psi} \left( 1 - g_3^x \right) + \left(\frac{\phi}{\psi}-h_4^x\right)
\end{gather}

\subsection{Solution of the fixed point equation}\label{app:fulleq}
The fixed-point equations as described in \eqref{eq:general_eq} for the given matrices $\tilde S, \eta(\tilde G), \tilde G$ is a priori a system of 
$26\times 26$ algebraic equations. are computed using Sympy in python, a symbolic calculation tool. In effect this is really a fixed point equation for $G$ a priori involving
$13\times 13$ algebraic equations. It turns out that many matrix elements vanish and (using the symbolic calculation tool Sympy in python)
we can extract a system of 39 algebraic equations which are given in the following:

\input{app-fullequation}

\subsection{Reduction of the solutions}
The previous system of equations can be reduced further by substitutions with a computer algebra system. We find the variables  $g_3^x, t_1^x, h_4^x, g_1^x, h_1^x$ are linked through the algebraic system:
\begin{equation}\label{eq:sys1}
    \left\{ \begin{array}{l}
    0 = 1 + g^{x}_{1} \left(- \mu^{2}  h^{x}_{4} - \frac{\phi}{\psi} g^{x}_{3} u^{2} +  x\right)\\
    0 = - h^{x}_{4} + g^{x}_{3} \left(- \mu^{2} \phi g^{x}_{1} h^{x}_{4} + \frac{\phi}{\psi} \right)\\
    0 =  \frac{\phi}{\psi} (1- g^{x}_{3}) -  g^{x}_{1} x - 1\\
    0 = \mu \psi g_1^x h_4^x - t_1^x\\
    0 = 1 - h_1^x - \mu t_1^x
    \end{array}\right.
\end{equation}
%\begin{equation}
%    \left\{ \begin{array}{l}
%        0 = \left(g^{x}_{1}\right)^{2} \nu^{2} x + g^{x}_{1} \left(- \mu^{2} h^{x}_{4} - \frac{\phi \nu^{2}}{\psi} + \nu^{2} + x\right) + 1\\
%        0 = - \frac{\mu^{2} \left(h^{x}_{1}\right)^{2}}{\psi} + h^{x}_{1} \left(- \mu^{2} g^{x}_{1} x - \mu^{2} + \frac{\mu^{2}}{\psi} - \frac{\nu^{2}}{\psi}\right) + \frac{\nu^{2}}{\psi}\\
%        0 = h^{x}_{1} \left(\frac{\phi}{\psi} - g^{x}_{1} x - 1\right) - h^{x}_{4}\\
%        0 = - \mu^{2} g^{x}_{1} h^{x}_{4} - \frac{h^{x}_{1}}{\psi} + \frac{1}{\psi}
%    \end{array}\right.
%\end{equation}
Notice this system can be shrinked further down to $3$ equations to get to the main result in \ref{th:main} using the substitution $h_1^x$ with the \nth{5} equation and $g_3^x$ with the \nth{3} equation. Also, by symmetry we find the same equations for $g_3^y, t_1^y, h_4^y, g_1^y, h_1^y$.

For the other variables, a set of equations link $q_1,q_2,q_4,q_5$. Notice there can many different representations depending on the reductions that are applied. Here we only show the example which has been used throughout the computations:
\begin{equation}\label{eq:sys2}
    \left\{\begin{array}{l}
        0 = - \mu^{2} g^{y}_{1} q_{2} + \mu^{2} h^{x}_{4} q_{1} + \mu g^{y}_{1} t^{x}_{1} + \mu g^{y}_{1} t^{y}_{1} - \frac{\phi g^{y}_{1} q_{4}\nu^{2}}{\psi} - g^{y}_{1} - q_{1} x + \frac{q_{1}\nu^{2} \left(\phi - \psi g^{x}_{1} x - \psi\right)}{\psi}\\
        0 = \mu \left(\phi - \psi g^{x}_{1} x - \psi\right) \left(- \mu g^{x}_{1} q_{2} + \mu h^{y}_{4} q_{1} + g^{x}_{1} t^{y}_{1}\right) + \frac{\phi h^{y}_{1} q_{4}}{\psi} - q_{2}\\
        0 = - \mu^{2} g^{x}_{1} h^{x}_{1} q_{4} + \frac{\mu^{2} q_{5} \left(\phi - \psi g^{y}_{1} y - \psi\right)}{\phi \psi} - g^{x}_{1} q_{4}\nu^{2} - q_{4} + \frac{q_{1}\nu^{2} \left(\phi - \psi g^{y}_{1} y - \psi\right)}{\phi}\\
        0 = \mu^{2} \phi g^{x}_{1} g^{y}_{1} h^{y}_{1} q_{4} - \frac{\mu^{2} g^{x}_{1} q_{5} \left(\phi - \psi g^{x}_{1} x - \psi\right)}{\psi} + \psi g^{x}_{1} g^{y}_{1} h^{y}_{1} + h^{y}_{1} q_{1} - \frac{q_{5}}{\psi}
    \end{array}\right.
\end{equation}

In conclusion, we can obtain 3 systems with $(4,5,5)$-equations or 3 systems with $(4,3,3)$-equations (so a total of 10), as in the main result \ref{th:main} (as discussed above these various systems are all equivalent and depend on the applied reductions).  

The solutions are not necessarily unique and one has to choose the appropriate ones with care. 
In our experimental results using Matlab with the "vpasolve" function, conditioning on $\Im g_1^x > 0$ and $\Im g_3^x > 0$ provided a unique solution to \eqref{eq:sys1} for $x \in \Real_+$ (or $x \in \Real \times i [0, \epsilon]$ for $\epsilon $ close to $0$ ); while conditioning on $g_1^x, g_3^x \in \Real_+$ provided a unique solution to \eqref{eq:sys1} for $x \in \Real_-$. We remind that we use $x \in \mathbb R_-$ exclusively in the time limit $t \to \infty$ in result \ref{result-infinite-time} while we use  $x \in \mathbb R_+$ in the situation of result \ref{th:main}. In addition, we found that selecting the appropriate solutions for $x$ and $y$ as just described for \eqref{eq:sys1} also led to a unique solution for \ref{eq:sys2} in our experiments.

%% file: app-fullequation.tex
\allowdisplaybreaks 
\begin{align}
0&=g^{x}_{1} \left(- \mu^{2} h^{x}_{4} + x\right) - g^{x}_{2} \nu + 1\\0&=g^{x}_{1} \left(- \mu^{2} h^{x}_{4} + x\right) + h^{x}_{2}\\0&=g^{x}_{2} \nu \left(- \mu^{2} h^{x}_{4} + x\right) + h^{x}_{5} \nu^{2}\\0&=- g^{y}_{1} \left(\mu^{2} q_{2} - \mu t^{x}_{1} - \mu t^{y}_{1} + 1\right) + \nu q^{xy}_{6} - q_{1} \left(- \mu^{2} h^{x}_{4} + x\right)\\0&=- g^{y}_{2} \nu \left(\mu^{2} q_{2} - \mu t^{x}_{1} - \mu t^{y}_{1} + 1\right) + \nu^{2} q_{3} - \nu q^{yx}_{6} \left(- \mu^{2} h^{x}_{4} + x\right)\\0&=g^{y}_{1} \left(\mu^{2} q_{2} - \mu t^{x}_{1} - \mu t^{y}_{1} + 1\right) - \nu q^{xy}_{6} + q_{1} \left(- \mu^{2} h^{x}_{4} + x\right)\\0&=\frac{\mu^{2} \phi g^{x}_{3} h^{x}_{3}}{\psi} - h^{x}_{1} + 1\\0&=\frac{\phi g^{x}_{3} h^{x}_{1}}{\psi} - h^{x}_{4}\\0&=- \frac{\mu \phi g^{x}_{3} h^{x}_{3}}{\psi} - t^{x}_{1}\\0&=\frac{\mu^{2} \phi g^{x}_{3} q_{5}}{\psi} + \frac{\mu^{2} \phi h^{y}_{3} q_{4}}{\psi} - \mu q^{yx}_{7}\\0&=\frac{\mu \phi g^{x}_{3} q^{xy}_{7}}{\psi} + \frac{\phi h^{y}_{1} q_{4}}{\psi} - q_{2}\\0&=- \frac{\phi g^{x}_{1} g^{x}_{3} \nu^{2}}{\psi} + g^{x}_{2} \nu\\0&=- \frac{\phi g^{x}_{1} g^{x}_{3} \nu^{2}}{\psi} - h^{x}_{2} + 1\\0&=\frac{\phi g^{x}_{3} h^{x}_{2} \nu^{2}}{\psi} - h^{x}_{5} \nu^{2}\\0&=- \frac{\phi g^{y}_{1} \nu^{2} q_{4}}{\psi} + \frac{\phi g^{x}_{3} \nu^{2} q_{1}}{\psi} - \nu q^{xy}_{6}\\0&=\frac{\phi g^{x}_{3} \nu^{3} q^{yx}_{6}}{\psi} + \frac{\phi h^{y}_{2} \nu^{2} q_{4}}{\psi} - \nu^{2} q_{3}\\0&=\frac{\phi g^{y}_{1} \nu^{2} q_{4}}{\psi} - \frac{\phi g^{x}_{3} \nu^{2} q_{1}}{\psi} + \nu q^{xy}_{6}\\0&=g^{x}_{3} \left(\frac{\mu^{2} h^{x}_{3}}{\psi} - g^{x}_{1} \nu^{2} - 1\right) + 1\\0&=g^{y}_{3} \left(\frac{\mu^{2} q_{5}}{\psi} + \nu^{2} q_{1}\right) + q_{4} \left(\frac{\mu^{2} h^{x}_{3}}{\psi} - g^{x}_{1} \nu^{2} - 1\right)\\0&=- \mu^{2} \psi g^{x}_{1} h^{x}_{1} - \mu^{2} h^{x}_{3}\\0&=- \mu^{2} \psi g^{x}_{1} h^{x}_{4} - h^{x}_{1} + 1\\0&=- \mu^{2} \psi g^{x}_{1} t^{x}_{1} + \mu \psi g^{x}_{1} + \mu h^{x}_{3}\\0&=- \mu^{3} \psi g^{x}_{1} q^{yx}_{7} - \mu^{2} \psi g^{x}_{1} h^{y}_{3} + \mu^{2} \psi h^{y}_{1} q_{1} - \mu^{2} q_{5}\\0&=- \mu^{2} \psi g^{x}_{1} q_{2} + \mu^{2} \psi h^{y}_{4} q_{1} + \mu \psi g^{x}_{1} t^{y}_{1} - \mu q^{xy}_{7}\\0&=- g^{x}_{2} \nu - h^{x}_{2} + 1\\0&=\mu \psi g^{y}_{1} h^{y}_{1} + \mu h^{y}_{3}\\0&=\mu \psi g^{y}_{1} h^{y}_{4} - t^{y}_{1}\\0&=- \frac{\phi g^{y}_{1} g^{y}_{3} \nu^{2}}{\psi} - h^{y}_{2} + 1\\0&=\frac{\phi g^{y}_{3} h^{y}_{2} \nu^{2}}{\psi} - h^{y}_{5} \nu^{2}\\0&=\frac{\phi g^{y}_{1} g^{y}_{3} \nu^{2}}{\psi} - g^{y}_{2} \nu\\0&=\frac{\mu^{2} \phi g^{y}_{3} h^{y}_{3}}{\psi} - h^{y}_{1} + 1\\0&=\frac{\phi g^{y}_{3} h^{y}_{1}}{\psi} - h^{y}_{4}\\0&=g^{y}_{3} \left(\frac{\mu^{2} h^{y}_{3}}{\psi} - g^{y}_{1} \nu^{2} - 1\right) + 1\\0&=- g^{y}_{2} \nu - h^{y}_{2} + 1\\0&=- \mu^{2} \psi g^{y}_{1} h^{y}_{1} - \mu^{2} h^{y}_{3}\\0&=- \mu^{2} \psi g^{y}_{1} h^{y}_{4} - h^{y}_{1} + 1\\0&=- g^{y}_{1} \left(- \mu^{2} h^{y}_{4} + y\right) - h^{y}_{2}\\0&=- g^{y}_{2} \nu \left(- \mu^{2} h^{y}_{4} + y\right) - h^{y}_{5} \nu^{2}\\0&=g^{y}_{1} \left(- \mu^{2} h^{y}_{4} + y\right) - g^{y}_{2} \nu + 1
\end{align}

%% file: app-replica.tex
\section{Linear pencil method from the replica trick argument}\label{app:replica}
A general approach to solve random matrix problems is to use the replica method, and historically this goes back to 
\cite{edwards1976eigenvalue}. In this appendix we show how to derive the fixed point equation 
\eqref{eq:general_eq} and \eqref{equ:sigmadef} as well as \eqref{equ:requirement} in appendix \ref{app:linear-pencil}. Such equations have been rigorously proved thanks to combinatorial methods in the recent literature on random matrix theory (see \cite{mingo2017free}, chapter 9, equ. (9.5)), but here we give a self-contained derivation using the replica trick, similar in spirit to \cite{edwards1976eigenvalue}. Although our derivation is far from rigorous it does covers linear pencils with a more general structure than in \cite{mingo2017free}, chapter 9, which are needed for our purposes.

\paragraph{Setting the replica calculation.} 
Let $(N_1, \ldots, N_d) \in \mathbb N^d$ for some $d \in \mathbb N$, and $N = \sum_{i=1}^d N_i$.
and let's consider a symmetric\footnote{Here $M$ plays the role of the symmetric matrix $\tilde M$ in appendix \ref{app:linear-pencil}. We remove the tilde to alleviate the notation as this will not create any confusion here} block matrix, called the "linear pencil", with $M = \sum_{i,j} E_{i,j} \otimes M^{(i,j)}$ such that $M^{(i,j)}$ is a matrix of size $N_i \times N_j$ and $E_{i,j}$ the matrix with elements $(E_{i,j})_{kl} = \delta_{ki} \delta_{lj}$. We assume that we can decompose $M = R + S$ with two block-matrices $R$ and $S$ such that the blocks $R^{(i,j)}$ are sum of independent real gaussian random matrices (with possibly their transpose) and $S^{(i,j)} = 0$ if $N_i\neq N_j$ or a scalar matrix $S^{(i,j)} = z_{i,j} \cdot I_{N_i}$ if $N_i=N_j$ (with $I_{N_i}$ the $N_i\times N_i$ identity).
Given the list of squares-blocks $\mathbb L = \{ i,j | N_i = N_j\}$, we let $B=(z_{i,j})$ be the matrix\footnote{This plays the role of $\tilde{B}$ in appendix \ref{app:linear-pencil}.} of the scalar coefficients in $\mathbb{C}$ where it is assumed $z_{i,j} = 0$ when $(i,j) \notin \mathbb L$.

Now we have defined a standard linear pencil. As a side remark, note also that we can accomodate random blocks $R^{(i,j)}$ which are {\it symmetric} gaussian random matrices (i.e., the lower and upper triangular parts are not independent) because we can always decompose them into the sum of two random matrices, $R^{(i,j)} = Y + Y^\intercal$.

In general, let $\{ Y_k \}_{1 \leq k \leq K}$ be a list of $K$ independent gaussian random matrices with i.i.d elements of variance $N^{-1}$, and various heights and widths among $\{N_i\times N_j$, $i, j \in \{1, \cdots, d\}\}$. These will constitute the random blocks of $R^{(i,j)}$ as follows. Given $i, j$ we define the set  $\mathbb S_{i,j} = \{ k | \text{width}(Y_k) = N_i, \text{height}(Y_k)= N_j\}$. We have for some coefficients $\alpha_{i,j}^{k}$, 
\begin{equation}
    R = \sum_{i,j = 1}^d 
    \sum_{k \in \mathbb S_{i,j}} \alpha^k_{i,j} (E_{i,j} \otimes Y_k
    + E_{j,i} \otimes Y_k^\intercal) .
\end{equation}
Notice $\alpha_{i,j}^{k}$ is not necessarily symmetric under exchange of $i$ and $j$, but $R = R^\intercal$ is still guaranteed to be symmetric (however, if $\alpha_{i,j}^{k}$ is symmetric for all $k$, then it implies all the random blocks are themselves symmetric).
Similarly, we have the scalar block,
\begin{equation}
    S = \sum_{(i,j) \in \mathbb L}^d z_{i,j} (E_{i,j} \otimes I_{N_i}) .
\end{equation}

Now, let's define $ I = \mean_{\{Y_k\}} \log \det(M)$. Then using the replica trick,
\begin{equation}
    I = -2 \mean \log \left[ \det (M)^{-\frac12} \right]
    \simeq 
    \lim_{n \to 0}
    2 \mean \left[ \frac{1 - \det (M)^{-\frac{n}{2}}}{n} \right] .
\end{equation}
We will first compute the term $J = \mean \det (M)^{-\frac{n}{2}}$ for integers $n\geq 1$. 

\paragraph{Calculation of $J$ for integer $n\geq 1$.}
We start with the Gaussian representation of the determinant
\begin{equation}
    J = \mean \left[ 
        \prod_{a=1}^n \int_{\mathbb R^N} \frac{\dd x}{(2 \pi)^{\frac{N}{2}}} 
        \exp\left( -\frac12 x^\intercal M x \right)
    \right] .
\end{equation}
We have
\begin{equation}
    J = \int_{\mathbb R^{n\times N}} 
    \exp \left( -\frac12 \sum_{a=1}^n x^{a \intercal} S x^a \right)
    \mean \left[ \exp \left( -\frac12 \sum_{a=1}^n x^{a \intercal} R x^a \right) \right] 
    \prod_{a=1}^n \frac{\dd x^a}{(2 \pi)^{\frac{N}{2}}}
\end{equation}
where $a=1, \dots, n$ is called the "replica index".
Setting $x^\intercal = [x_1 | \ldots | x_d]$ where each $x_i$ is of size $N_i$,
\begin{equation}
    x^\intercal R x 
    = \sum_{i,j=1}^d \left[
    x_i^\intercal \left( \sum_{k \in \mathbb S_{i,j}} \alpha_{i,j}^k Y_k \right) x_j
    + x_j^\intercal \left( \sum_{k \in \mathbb S_{i,j}} \alpha_{i,j}^k Y_k^\intercal \right) x_i
    \right] .
\end{equation}
Then defining the set $\mathbb S^{-1}_k = \{ (i,j) | N_i = \text{width}(Y_k), N_j = \text{height}(Y_k)\}$ (for a given $k$)
\begin{equation}
    x^\intercal R x 
    = \sum_{k=1}^{K} \sum_{(i,j) \in \mathbb S_k^{-1}} 2 \alpha_{i,j}^k (x_i^\intercal Y_k x_j)
\end{equation}
an expanding the inner products we can further write down
\begin{equation}
    x^\intercal R x 
    = 2 \sum_{k=1}^K \sum_{r,s} [Y_k]_{r,s} \sum_{(i,j) \in \mathbb S_k^{-1}} \alpha_{i,j}^k [x_i]_r [x_j]_s
\end{equation}
Thus
\begin{equation}
    \mean \left[ \exp \left( -\frac12 \sum_{a=1}^n x^{a \intercal} R x^a \right) \right]
    = \prod_{k,r,s} \mean \left[
        \exp \left( -
            [Y_k]_{r,s} \sum_{\substack{1 \leq a \leq n\\(i,j) \in \mathbb S_k^{-1}}} \alpha_{i,j}^k [x_i^a]_r [x_j^a]_s
        \right)
    \right]
\end{equation}
and using the moment generating function of the normal distribution, with $\mathbb{E} [Y_k]_{r,s}^2 = \frac{1}{N}$, we obtain
\begin{equation}
    \mean \left[ \exp \left( -\frac12 \sum_{a=1}^n x^{a \intercal} R x^a \right) \right]
    = 
    \exp \left( 
        \frac{1}{2N} \sum_{k,r,s}
        \left( \sum_{\substack{1 \leq a \leq n\\(i,j) \in \mathbb S_k^{-1}}} \alpha_{i,j}^k [x_i^a]_r [x_j^a]_s \right)^2 
    \right).
\end{equation}
But now, we can expand the square using the set $\mathbb T = \{(i,j,k,l) \in \{1, \ldots, d\}^4 | (N_i,N_j) = (N_k,N_l)\}$:
\begin{equation}
    \sum_{k,r,s}
    \left(\sum_{\substack{1 \leq a \leq n\\(i,j) \in \mathbb S_k^{-1}}} \alpha_{i,j}^k [x_i^a]_r [x_j^a]_s \right)^2 
    = \sum_{\substack{1 \leq a \leq n\\1 \leq b \leq n}}
    \sum_{\substack{i_a,j_a,i_b,j_b\\\in \mathbb T}}
    (x_{i_a}^a \cdot x_{i_b}^b)
    (x_{j_a}^a \cdot x_{j_b}^b)
    \sum_{k \in \mathbb S_{i_a,j_a}} \alpha_{i_a,j_a}^k \alpha_{i_b,j_b}^k .
\end{equation}
% ATTENTION: je crois finalement que c'est nécessaire de faire cette remarque: OUI (N)
Notice the symmetry with the indices
\begin{equation}
    \sum_{\substack{i_a,j_a,i_b,j_b\\\in \mathbb T}}
    (x_{i_a}^a \cdot x_{i_b}^b)
    (x_{j_a}^a \cdot x_{j_b}^b)
    \sum_{k \in \mathbb S_{i_a,j_a}} \alpha_{i_a,j_a}^k \alpha_{i_b,j_b}^k
    = 
    \sum_{\substack{j_a,i_a,j_b,i_b\\\in \mathbb T}}
    (x_{i_a}^a \cdot x_{i_b}^b)
    (x_{j_a}^a \cdot x_{j_b}^b)
    \sum_{k \in \mathbb S_{j_a,i_a}} \alpha_{j_a,i_a}^k \alpha_{j_b,i_b}^k .
\end{equation}
% ATTENTION: je crois finalement que c'est nécessaire de faire cette definition car sinon on n'aurait pas \sigma_{i,j}^{l,k} =\sigma_{j,i}^{k,l} !!! OUI (N)
Therefore, defining $\sigma_{i,j}^{l,k} \equiv \sum_{t \in \mathbb S_{i,j}} \alpha_{i,j}^t \alpha_{k,l}^t + \sum_{t \in \mathbb S_{j,i}} \alpha_{j,i}^t \alpha_{l,k}^t$
%Therefore, with $\sigma_{i,j}^{l,k} \equiv 2 \sum_{t \in \mathbb S_{i,j}} \alpha_{i,j}^t \alpha_{k,l}^t$ we write:
\begin{equation}
    \sum_{\substack{i_a,j_a,i_b,j_b\\\in \mathbb T}}
    (x_{i_a}^a \cdot x_{i_b}^b)
    (x_{j_a}^a \cdot x_{j_b}^b)
    \sum_{k \in \mathbb S_{i_a,j_a}} \alpha_{i_a,j_a}^k \alpha_{i_b,j_b}^k
    = \frac{1}{2} \sum_{\substack{i_a,j_a,i_b,j_b\\\in \mathbb T}}
    (x_{i_a}^a \cdot x_{i_b}^b)
    (x_{j_a}^a \cdot x_{j_b}^b) \sigma_{i_a,j_a}^{j_b,i_b} .
\end{equation}
We remark for further use the symmetry property $\sigma_{i,j}^{l,k} = \sigma_{j,i}^{k,l}$.

Now, notice also that we have $x^\intercal B x = \sum_{(i,j)\in \mathbb L} z_{i,j} (x_i \cdot x_j)$, so
\begin{equation}
    J = \int_{\mathbb R^{n\times N}} \!\!
    \exp \biggl(
        -\frac12 \sum_{\substack{1 \leq a\leq n\\(i,j)\in \mathbb L}} z_{i,j} (x_i^a \cdot x_j^a)
        + \frac{1}{4N} \sum_{\substack{1 \leq a \leq n\\1 \leq b \leq n}}  \sum_{\substack{i_a,j_a,i_b,j_b \\ \in \mathbb T}}
        (x_{i_a}^a \cdot x_{i_b}^b)
        (x_{j_a}^a \cdot x_{j_b}^b) \sigma_{i_a,j_a}^{j_b,i_b}
    \biggr) \!\!
    \prod_{a=1}^n \frac{\dd x^a}{(2 \pi)^{\frac{N}{2}}}
\end{equation}
Now, let's define the "overlaps" $q_{i,j}^{a,b} = \frac{1}{N} x^a_i \cdot x^b_j$ for $i,j \in \mathbb L$ and $1 \leq a,b \leq n$, and $0$ otherwise, then
\begin{equation}
    J = \int_q \int_{\mathbb R^{n\times N}} 
    \left( \prod_{a=1}^n \frac{\dd x^a}{(2 \pi)^{\frac{N}{2}}} \right)
    \left( \prod_{ \substack{(i,j) \in \mathbb L\\1 \leq a \leq b \leq n} } \dd q^{a,b}_{i,j} 
        \delta \left( q^{a,b}_{i,j} - \frac{x_i^a \cdot x_i^b}{N} \right)
    \right)
    e^{N\Xi (q)}
\end{equation}
with
\begin{equation}
    \Xi(q) =
    -\frac{1}{2} \sum_{a=1}^n \sum_{(i,j) \in \mathbb L} z_{i,j} q_{i,j}^{a,a}
    + \frac14 \sum_{\substack{1 \leq a \leq n\\1 \leq b\leq n}}  \sum_{\substack{i_a,j_a,i_b,j_b\\ \in \mathbb T}}
    q_{i_a, i_b}^{a,b} q_{j_a, j_b}^{a,b} \sigma_{i_a,j_a}^{j_b,i_b} .
\end{equation}
Eventually, with a Fourier transform representation of the Dirac  distribution and a change of variable to have real integrands
\begin{equation}
    J = \int_{\hat q} \int_q \int_{\mathbb R^{n\times N}} 
    \left( \prod_{a=1}^n \frac{\dd x^a}{(2 \pi )^{\frac{N}{2}}} \right)
    \left( 
        \prod_{ \substack{(i,j) \in \mathbb L\\1 \leq a \leq b \leq n} }
        \dd q^{a,b}_{i,j} \dd \hat q^{a,b}_{i,j} 
        \frac{N}{2 \pi i}
        e^{- \hat q^{a,b}_{i,j} (N q^{a,b}_{i,j} - x^a_i \cdot x_j^b)}
    \right)
    e^{N\Xi (q)} .
\end{equation}
So for some constant $C$ we have (the constant turns out to be unimportant in the high-dimensional limit)
\begin{equation}
    J = C \int_{\hat q} \int_q 
    \left( \prod_{ \substack{i,j\\1 \leq a \leq b \leq n} }^n \dd q^{a,b}_{i,j} \dd \hat q^{a,b}_{i,j} \right)
    e^{N (\Gamma(q, \hat q) + \Xi(q)) + \psi(\hat q))}
\end{equation}
where
\begin{equation}
    \Gamma(q,\hat q) = -\sum_{(i,j) \in \mathbb L} \sum_{1 \leq a \leq b \leq n} \hat q_{i,j}^{a,b} q_{i,j}^{a,b}
\end{equation}
and
\begin{equation}
    \psi(\hat q) = 
    \frac{1}{N}
    \log \int_{\mathbb R^{n\times N}} 
    \left( \prod_{\substack{1\leq a \leq n\\1 \leq k \leq d}} \frac{\dd x^a_k}{(2 \pi)^{\frac{N_k}{2}}} \right)
    e^{\sum_{(i,j) \in \mathbb L} \sum_{1 \leq a \leq b \leq n} \hat q_{i,j}^{a,b} (x_i^a  \cdot x_{j}^b)} .
\end{equation}
Notice that for $\psi(\hat q)$ we can further expand the terms over components of $x_k^a = [x_k^a]_{r=1}^{N_k}$
\begin{equation}
    \psi(\hat q) = 
    \frac{1}{N}
    \log \int_{\mathbb R^{n\times N}} 
    \left( \prod_{\substack{1\leq a \leq n\\1 \leq k \leq d}} 
    \prod_{r=1}^{N_k}
    \frac{\dd [x^a_k]_r}{\sqrt{2 \pi}} \right)
    \exp \left( \sum_{\substack{(i,j) \in \mathbb L\\1 \leq a \leq b \leq n}} \sum_{r=1}^{N_i} \hat q_{i,j}^{a,b} [x_i^a]_r [x_{j}^b]_r \right) .
\end{equation}
Now setting $\phi(q,\hat q) = \Gamma(q,\hat q) + \Xi(q) + \psi(\hat q)$, the saddle point method provides for $N$ large enough
\begin{equation}
    \frac{\log{J}}{N} \simeq \text{Extr}(\phi(q,\hat q)) .
\end{equation}

\paragraph{Replica symmetric ansatz.}
Before computing the extremum we make the "replica symmteric ansatz": we assume for all $(i,j)$, $q^{a,b}_{i,j} = q_{i,j} \delta_{a,b} \delta_{(i,j) \in \mathbb L}$ and $\hat q^{a,b}_{i,j} = -\frac12 \hat q_{i,j} \delta_{a,b} \delta_{(i,j) \in \mathbb L}$. As shown here with this ansatz 
$\phi(q,\hat q)$ will become tractable.
We have
\begin{equation}
    \Gamma(q, \hat q) = \frac{n}{2} \sum_{(i,j) \in \mathbb L} q_{i,j} \hat q_{i,j} .
\end{equation}
Furthermore, we can calculate $\psi(\hat q)$ noticing that
\begin{equation}
    \psi(\hat q) =
    \frac{1}{N}
    \log \int_{\mathbb R^{n\times N}} 
    \left( \prod_{\substack{1\leq a \leq n\\1 \leq k \leq d}} 
    \prod_{r=1}^{N_k}
    \frac{\dd [x^a_k]_r}{\sqrt{2 \pi}} \right)
    \exp \left( -\frac12 \sum_{(i,j) \in \mathbb L}
    \hat q_{i,j} 
    \sum_{r=1}^{N_i} \sum_{1 \leq a \leq n} [x_i^a]_r [x_{j}^a]_r \right) ,
\end{equation}
so
\begin{equation}
    \psi(\hat q) =
    \frac{n}{N}
    \log \int_{\mathbb R^{N}} 
    \left( \prod_{1 \leq k \leq d}
    \prod_{r=1}^{N_k}
    \frac{\dd [x_k]_r}{\sqrt{2 \pi}} \right)
    \exp \left( -\frac12 \sum_{(i,j) \in \mathbb L}
    \hat q_{i,j} 
    \sum_{r=1}^{N_i} [x_i]_r [x_{j}]_r \right) .
\end{equation}
But notice also that we can group the terms. Defining the "equivalence class of $i$"  as the set  $\bar i = \{j | N_i = N_j\}$, we get
\begin{equation}
    \psi(\hat q) =
    \frac{n}{N}
    \log \int_{\mathbb R^{N}} 
    \prod_{1 \leq i \leq d}  \prod_{r=1}^{N_i} \left(
    \frac{\dd [x_i]_r}{\sqrt{2 \pi}} 
    \prod_{j \in \bar i}
    e^{ -\frac12 
    \hat q_{i,j} [x_i]_r [x_{j}]_r 
    } \right) .
\end{equation}
But now, since the equivalence classes forms a partition $\mathcal P$ of $\{1, \ldots, d\}$, we have
\begin{equation}
    \psi(\hat q) =
    \frac{n}{N}
    \log \int_{\mathbb R^{N}} 
    \prod_{\bar k \in \mathcal P}
    \prod_{i \in \bar k}  \prod_{r=1}^{N_k} \left(
    \frac{\dd [x_i]_r}{\sqrt{2 \pi}} 
    \prod_{j \in \bar k}
    e^{ -\frac12 
    \hat q_{i,j} [x_i]_r [x_{j}]_r 
    } \right) .
\end{equation}
Hence
\begin{equation}
    \psi(\hat q) =
    \frac{n}{N}
    \log \prod_{\bar k \in \mathcal P} \left[ \int_{\mathbb R^{|\bar k|}} 
    \prod_{i \in \bar k}   \left(
    \frac{\dd y_i}{\sqrt{2 \pi}} 
    \prod_{j \in \bar k}
    e^{ -\frac12 
    \hat q_{i,j} y_i y_{j}
    } \right)
    \right]^{N_k}
\end{equation}
Or written in a slightly different way
\begin{equation}
    \psi(\hat q) =
    n
    \sum_{\bar k \in \mathcal P} \frac{N_k}{N}
    \log \left[ \int_{\mathbb R^{|\bar k|}} 
    \left( \prod_{i \in \bar k} 
    \frac{\dd y_i}{\sqrt{2 \pi}} \right)
    e^{ -\frac12  \sum_{(i,j) \in \bar k}
    \hat q_{i,j} y_i y_{j}
    } 
    \right] .
\end{equation}
We define the overlap matrix $\hat Q = (\hat q_{i,j})$ and the sub-matrix $\hat Q^{\bar k} = (\hat Q_{i,j})_{(i,j) \in \bar k}$. 
Recalling that for a multivariate gaussian distribution
\begin{equation}
    \int_{\mathbb R^{|\bar k|}} 
    \left( \prod_{i \in \bar k} 
    \frac{\dd y_i}{\sqrt{2 \pi}} \right)
    e^{ -\frac12  \sum_{(i,j) \in \bar k}
    \hat q_{i,j} y_i y_{j}
    } 
    = 
    \left(\det \hat Q^{\bar k} \right)^{-\frac12}
\end{equation}
we find
\begin{equation}
    \psi(\hat q) = -\frac{n}{2} \sum_{\bar k \in \mathcal P} \frac{N_k}{N}
    \log \det \hat Q^{\bar k}
\end{equation}
Finally, for the term $\Xi(q)$ we obtain
\begin{equation}
    \Xi(q) = - \frac{n}{2} \sum_{(i,j) \in \mathbb L} z_{i,j} q_{i,j} 
    + \frac{n}{4} \sum_{\substack{(i,j,k,l) \in \mathbb T}}
    q_{i, k} q_{j, l} \sigma_{i,j}^{l,k} .
\end{equation}
Summarizing, we have found
\begin{align}
    \phi(q, \hat q)
    & = \frac{n}{2}\Bigg\{
    - \sum_{(i,j) \in \mathbb L} z_{i,j} q_{i,j} 
    + \frac12 \sum_{\substack{i,j,k,l\\ \in \mathbb T}}
    q_{i, k} q_{j, l} \sigma_{i,j}^{l,k}
    + \sum_{(i,j)  \in \mathbb L} q_{i,j} \hat q_{i,j}
    - \sum_{\bar k \in \mathcal P} \frac{N_k}{N}
    \log \det \hat Q^{\bar k} \Biggr\}
    \nonumber \\ &
    \equiv 
    \frac{n}{2} \tilde{\phi}(q, \hat q) .
\end{align}

\paragraph{Derivation of fixed point equation \eqref{eq:general_eq}.}
Now we will have to take derivatives to find the extremum of $\phi(q,\hat q)$ or equivalently $\tilde{\phi}(q, \hat q)$. In order to perform the derivatives it is useful to recall that for a symmetric matrix $X$ we have
\begin{equation}
    \frac{\partial \log \det X}{\partial [X]_{i,j}} 
    = \frac{1}{\det X} \frac{\partial \det X}{\partial [X]_{i,j}} 
    = [X^{-1}]_{ji} = [X^{-1}]_{i,j} .
\end{equation}
Therefore, we have for any $(i,j) \in \mathbb L$ (using the symmetry of $\sigma_{i,j}^{l,k}$)
\begin{gather}
    \frac{\partial \tilde{\phi}(q,\hat q)}{\partial \hat q_{i,j}} = 0 
    \Longrightarrow q_{i,j} = \frac{N_i}{N} [(\hat Q^{\bar i})^{-1}]_{i,j}
    \\
    \frac{\partial \tilde{\phi}(q,\hat q)}{\partial q_{i,j}} = 0 
    \Longrightarrow 
    z_{i,j} = \hat q_{i,j} + \sum_{(k,l) \in \mathbb L} q_{k,l} \sigma_{i,k}^{l,j}
\end{gather}
In matrix form, using the matrix $G$ with matrix elements $G_{i,j} = \frac{N}{N_i}Q_{i,j}$, and given an equivalence class $\bar k$ we have
\begin{gather}
    Q^{\bar k} = \frac{N_k}{N} (\hat Q^{\bar k})^{-1}\\
    B^{\bar k} = \hat Q^{\bar k} + \eta^{\bar k}\left(G \right)
\end{gather}
where for any given matrix $D\in \mathbb R^{d\times{d}}$, the matrices $B^{\bar k}$, $\eta^{\bar k}(D)$ are  the restriction of 
$B$, $\eta(D)$ on the subspace spanned by the basis $\mathcal B^{\bar k}$ (that is, on all the indices $(i,j) \in \bar k\times \bar k$), and $\eta(D)$ is defined such that for any $(i,j) \in \mathbb L$
\begin{equation}
    \left[\eta(D) \right]_{i,j} =
    \sum_{(u,v) \in \mathbb L} \frac{N_u}{N} [D]_{u,v} \sigma_{i,u}^{v,j} .
\end{equation}
So for any $k \in \{1, \ldots, d\}$ we have
\begin{equation}
    B^{\bar k} =  \left( \frac{N}{N_k} Q^{\bar k}\right)^{-1} + \eta^{\bar k} \left( G \right)
\end{equation}
Hence using only $G$, because $G^{\bar k} = \frac{N}{N_k} Q^{\bar k}$, we obtain
\begin{equation}\label{eq:fixed_point_eq1}
    B^{\bar k} G^{\bar k} = I_{|\bar k|} + \eta^{\bar k}(G) G^{\bar k} .
\end{equation}

Notice now that $X \in \mathbb R^{|\bar k|} \mapsto \hat Q^{\bar k} X$ is related to the endomorphism restriction of $X \mapsto \hat Q X$ in the vector space spanned by the canonical basis $\mathcal B^{\bar k} = (e_{i})_{i \in \bar k}$. In other words, we have that $\mathbb R^d = \bigoplus_{\bar k \in \mathcal P} \mathcal B^{\bar k}$, and that the subspaces $\mathcal B^{\bar k}$ are stable under action of $\hat Q$, but also under action of $B$ as there is also the constraint that $\forall (i,j) \notin \mathbb L, z_{i,j} = 0$. Similarly this is the case also for $\eta$, since by definition of $\eta$, we have $[\eta(D)]_{i,j} = 0$ for any $(i,j) \notin \mathbb L$. In other words, assuming the partition formed by the equivalence classes $\bar i = \{j | N_j =N_i\}$ is $\mathcal P = \{\bar k_1, \ldots, \bar k_p \}$, there exists a matrix $P \in \mathbb R^{d \times d}$ such that
\begin{equation}
    P^{-1} B P = \left( \begin{array}{c|c|c|c}
        B^{\bar{k_1}} & 0 & \hdots & 0\\ \hline
        0 & B^{\bar{k_2}} & \hdots & 0\\ \hline
        \vdots & \vdots & \ddots & \vdots\\ \hline
        0 & 0 & \hdots & B^{\bar{k_p}}\\
    \end{array} \right),  ~~~
    P^{-1} G P = \left( \begin{array}{c|c|c|c}
        G^{\bar{k_1}} & 0 & \hdots & 0\\ \hline
        0 & G^{\bar{k_2}} & \hdots & 0\\ \hline
        \vdots & \vdots & \ddots & \vdots\\ \hline
        0 & 0 & \hdots & G^{\bar{k_p}}\\
    \end{array} \right)
\end{equation}
and similarly with the same matrix $P$
\begin{equation}
    P^{-1} \eta(D) P = \left( \begin{array}{c|c|c|c}
        \eta^{\bar{k_1}}(D) & 0 & \hdots & 0\\ \hline
        0 & \eta^{\bar{k_1}}(D)  & \hdots & 0\\ \hline
        \vdots & \vdots & \ddots & \vdots\\ \hline
        0 & 0 & \hdots & \eta^{\bar{k_p}}(D)\\
    \end{array} \right)
\end{equation}
Therefore, since we have for all $\bar k \in \mathcal P$ the equation \eqref{eq:fixed_point_eq1}, this is equivalent to having $(P^{-1}BP) (P^{-1}GP) = I_d + (P^{-1} \eta(D) P) (P^{-1}GP)$, in other words, this is equivalent to
\begin{equation}
    B G = I_d + \eta(G) G
\end{equation}
At this point we have derived the important fixed point equation \eqref{eq:general_eq}. 

\paragraph{Derivation of equation \eqref{equ:requirement}.}
Notice that we also have (because $M$ is symmetric)
\begin{equation}
    \frac{\partial I}{\partial z_{i,j}} 
    = \sum_{r,s=1}^N \mean \frac{\partial [M]_{r,s}}{\partial z_{i,j}} \frac{\partial \log \det M}{\partial [M]_{r,s}}
    = \sum_{r,s=1}^N \mean \frac{\partial [S]_{r,s}}{\partial z_{i,j}} [M^{-1}]_{r,s}
    = \mean \trace{(M^{-1})^{(i,j)}}
\end{equation}
But on the other hand, with $(q^*,\hat q^*)$ the extrema of $\tilde{\phi}$
\begin{equation}
    I = 2 \lim_{n\to 0}\frac{1 - \mathbb{E}J}{n}
    \simeq
    2 \lim_{n\to 0} \frac{1-e^{- N \frac{n}{2} \tilde{\phi}(q^*, \hat q^*)}}{n}
    =  -N \tilde{\phi}(q^*, \hat q^*) .
\end{equation}
But notice that $\phi$ and $q,q^*$ are also themselves functions of $B$, in other words
\begin{equation}
    \frac{I}{N} \simeq - \tilde{\phi}(q^*(B), \hat q^*(B), B) ,
\end{equation}
and hence using chain rule, and remembering that $\frac{\partial \tilde{\phi}}{\partial q_{i,j}} = \frac{\partial \tilde{\phi}}{\partial \hat q_{i,j}} = 0$ in $q^*,\hat q^*$, we have
\begin{equation}
    \frac{1}{N} \frac{\partial I}{\partial z_{i,j}} 
    = - \sum_{r,s=1}^d \frac{\partial [B]_{r,s}}{\partial z_{i,j}} \frac{\partial \tilde{\phi}}{\partial [B]_{r,s}} (q^*, \hat q^*, B) + 0 + 0
    = q_{i,j}^* = \frac{N_i}{N} [G^*]_{i,j} .
\end{equation}
Eventually we obtain
\begin{equation}
    [G^*]_{i,j} = \lim_{N \to \infty} \frac{1}{N_i} \mean \trace{(M^{-1})^{(i,j)}}
\end{equation}
which is \eqref{equ:requirement}.

\paragraph{Derivation of equation \eqref{equ:sigmadef}.}
Regarding $\sigma_{i,j}^{l,k}$, notice that we can express the random block $R^{(i,j)}$ in the following way
\begin{equation}
    R^{(i,j)} = \sum_{t \in \mathbb S_{(i,j)}} \alpha^t_{i,j} Y_t + \sum_{t \in \mathbb S_{(j,i)}}\alpha^t_{j,i} Y_t^\intercal
\end{equation}
so, provided $r,s$ are chosen such that $r\neq s$, we find
\begin{equation}\label{eq:randomblocks_sigma}
    N \cdot \mean \left[ [R^{(i,j)}]_{r,s} [R^{(l,k)}]_{s,r} \right] = 
    \sum_{t\in \mathbb S_{(i,j)}} \alpha_{i,j}^t \alpha_{k,l}^t 
    + \sum_{t\in \mathbb S_{(j,i)}} \alpha_{j,i}^t \alpha_{l,k}^t
    = \sigma_{i,j}^{l,k} 
\end{equation}
which is nothing else than \eqref{equ:sigmadef}.

\paragraph{A note on correlated random matrices.}
To extend further the result, notice that we can always construct standard gaussian random blocks, say  $R^{(i,j)}$ and $R^{(l,k)}$, such that they have a priori some covariance $v$ with $v = \mean \left[ [R^{(i,j)}]_{r,s} [R^{(l,k)}]_{s,r} \right] ]$. While we stated a result where these blocks are built from a sum of $(Y_k)$ which are standard gaussian random matrices, notice that it is always possible to use two independent standard random matrices $Y_1,Y_2$, and define: $R^{(i,j)} = v Y_1 + \sqrt{1-v^2} Y_2$ and $R^{(k,l)} =  Y_1^T $. Therefore, the result remains valid even in the general case where we only suppose that the blocks in $R$ are distributed following a gaussian distribution, with potentially some entry-wise covariance and using equation \eqref{eq:randomblocks_sigma} as the definition of $\sigma_{i,j}^{l,k}$.

%% file: app-numerical-experiments.tex
\section{Numerical results}\label{app:numerical-experiments}
All the experiments are run on a standard desktop configuration:
\begin{enumerate}
    \item Matlab R2019b is used to generate the heatmaps or 3D landscapes. Most exemples can be generated in less than 12h on a standard machine.
    \item The experimental comparisons run on a standard instance of a Google collaboratory notebook in less than a few hours.
\end{enumerate}

\subsection{Numerical computations}
We take equation \eqref{eq:repr_g} as an example of how to proceed with the numerical experiments. Specifically we consider the second integral in the Cauchy integral representation of $g(t)$
\begin{equation}
    g_2(t) = -\frac{1}{2i\pi} \oint_{\Gamma} \dd z \left\{ 
        \frac{ 1-e^{-t(z+\delta)} }{z+\delta} K(z) 
    \right\}.
\end{equation}
We choose a contour with $\lambda^* = \max \Sp \frac{Z^TZ}{N}$ with two positive fixed constants $\epsilon, \Delta$:
\begin{equation}
    \Gamma = 
    \{ \gamma \lambda^* \pm i\Delta | -\epsilon \leq \gamma \leq 1 +\epsilon \}
    \cup
    \{ \epsilon \lambda^* + \gamma i\Delta | -1 \leq \gamma \leq 1 \}
    \cup
    \{ -\epsilon + \gamma i\Delta | -1 \leq \gamma \leq 1 \}
\end{equation}
Now, the integrand is continuous in $\lambda^\star+\epsilon$ and $-\epsilon$ for $\epsilon$ small enough. So taking the limit $\epsilon \to 0$ and $\Delta \to 0$
\begin{equation}
    g(t) = \lim_{\Delta \to 0} \frac{1}{2i\pi} 
    \int_0^{\lambda^*} \biggl\{ 
        \frac{ 1-e^{-t(r+\delta + i\Delta)} }{r+\delta + i\Delta} K(r + i\Delta)
        - \frac{ 1-e^{-t(r+\delta) - i\Delta} }{r+\delta - i\Delta} K(r - i \Delta)
    \biggr\} d r
\end{equation}
which is simply
\begin{equation}\label{eq:gcompute}
    g(t) = 
    \int_0^{\lambda^*}
    \frac{ 1-e^{-t(r+\delta)} }{r+\delta}
    \lim_{\Delta \to 0} 
    \frac{1}{2i\pi} 
    \biggl\{ 
         K(r + i\Delta) - K(r - i\Delta) 
    \biggr\} d r
\end{equation}
Obviously the inward term is also given by the limit $\lim_{\Delta \to 0} \frac{1}{\pi} \Im K(r+i\Delta)$.
% as $\Im K(r+i\Delta) = \frac{1}{2i} (K(r+i\Delta) - K(r-i\Delta))$.
So this all there is to compute from the former algebraic equations are appropriate imaginary parts. This can be done by taking a discretized interval $0\leq r_1 \leq \ldots \leq r_K\leq \lambda^*$, and solving the algebraic equations for the imaginary value $\Im t_1^x$ for $x=r_i$, $i=1,\cdots, K$.
%$\Im g_1^x, \Im h_4^x, \ldots$.

We proceed similarly with the terms containing two complex variables $x$ and $y$ (or two resolvents). For instance for $W(x,y)$ one uses the limit in $\Delta_x, \Delta_y \to 0$ of $\rho(x,y)$ where
\begin{align}
    \rho(x,y)  =  & \lim_{\Delta_x \to 0} \lim_{\Delta_y \to 0} 
    \biggl[
     \frac{-1}{4 \pi^2}  \biggl\{
    W(r_x + i\Delta_x, r_y + i\Delta_y) - 
    W(r_x + i\Delta_x, r_y - i\Delta_y) \biggr\}
    \nonumber \\ &
    - 
    \frac{-1}{4 \pi^2} \biggl\{
    W(r_x - i\Delta_x, r_y + i\Delta_y) - 
    W(r_x - i\Delta_x, r_y - i\Delta_y) \biggr\}
    \biggr]
\end{align}
or equivalenlty
\begin{align}
    \rho(x,y) = \lim_{\Delta_x, \Delta_y \to 0}
    \frac{1}{2 \pi^2} \Re \biggl\{ W(r_x + i\Delta_x, r_y - i\Delta_y) - W(r_x + i\Delta_x, r_y + i\Delta_y) \biggr\}
\end{align}

\subsection{Technical considerations}

\paragraph{Dirac distributions with 1-variable functions:}
It happens that the limiting distribution $\frac{Z^TZ}{N}$ may contain a mixture of a Dirac peak at $0$ and a continuous measure. For instance, $K(z)$ may contain a branch cut in the interval $\mathcal C^* = [\lambda_1, \lambda^*]$ with $\lambda_0=0<\lambda_1<\lambda^*<\infty$ along with an isolated pole in $0$ with: $K(z) =\frac{\alpha}{0-z} + K_c(z)$ (where $ K_c : \Complex \setminus \mathcal C^* \to \Complex$). For instance, equation \eqref{eq:gcompute} becomes:
\begin{equation}
    g(t) = \alpha \frac{1-e^{-t \delta}}{\delta} + \int_{\lambda_0}^{\lambda^*} \dd r \frac{1-e^{-t(\delta +r)}}{r+\delta} \lim_{\Delta \to 0} \frac{1}{\pi} \Im K_c(r+i\Delta) 
\end{equation}
The weight $\alpha$ can be retrieved by computing $\alpha = \lim_{\epsilon \to 0^+} (-i\epsilon) K(i\epsilon) = \lim_{\epsilon \to 0^+} \epsilon \Im K(i\epsilon)$.

\paragraph{Dirac distributions with 2-variables functions:}
Similarly, we can have an isolated pole at $0$ for $x,y$ for $W(x,y)$. In that case, we can write down $W(x,y)$ as for instance:
\begin{equation}
    W(x,y) = \frac{\alpha_{xy}}{(0-x)(0-y)} + \frac{\alpha_x}{0-x} W_y(y) + \frac{\alpha_y}{0-y} W_x(x) + W_{xy}(x,y)
\end{equation}
where $W_x,W_y$ are defined on $\Complex \setminus \mathcal C^* \to \Complex$ and $W_{xy}: (\Complex \setminus \mathcal C^*)^2 \to \Complex$. Firstly, We can easily find $\alpha_{xy}$ with:
\begin{equation}
    \alpha_{xy} = \lim_{\epsilon \to 0^+} (- \epsilon^2) \Re W(i \epsilon, i \epsilon)
\end{equation}
Secondly, all the considered 2-variables functions are symmetrical with respect to $x$ and $y$: $W(x,y) = W(y,x)$ which implies that $\alpha_x = \alpha_y$ and $W_x(r) = W_y(r)$ for all $r \in \Complex \setminus \mathcal C^*$.
Therefore, if we have $\gamma_t(z) = \frac{1-e^{-t(\delta+z)}}{z+\delta}$, we have to compute:
\begin{equation}
    \begin{split}
        \opR_{x,y} \left\{ \gamma_t(x)\gamma_t(y) W(x,y)\right\}
        = \gamma_t(0)^2 \alpha_{xy}
        + \iint_{[\lambda_0, \lambda^*]^2}  \gamma_t(u) \gamma_t(v) \rho(u,v) \dd u \dd v\\
        +  2 \gamma_t(0) \int_{\lambda_0}^{\lambda^*} \dd r \gamma_t(r) 
        \lim_{\Delta \to 0^+} \frac{\alpha_x}{2 i \pi} \biggl\{W_y( r + i \Delta) - W_y(r - i \Delta)\biggr\}
\end{split}
\end{equation}
But because we don't have access to $\alpha_x$ nor $W_y$ directly, we can use the full form:
\begin{equation}
    \begin{split}
        \opR_{x,y} \left\{ \gamma_t(x)\gamma_t(y) W(x,y)\right\}
        = \gamma_t(0)^2 \alpha_{xy}
        + \iint_{[\lambda_0, \lambda^*]^2}  \gamma_t(u) \gamma_t(v) \rho(u,v) \dd u \dd v\\
        +  2 \gamma_t(0) \int_{\lambda_0}^{\lambda^*} \dd r \gamma_t(r) 
        \lim_{\Delta \to 0^+} \lim_{\epsilon \to 0^+}  \frac{-i \epsilon}{2 i \pi} \biggl\{ W(i \epsilon, r + i \Delta) - W(i \epsilon, r - i \Delta)\biggr\}
\end{split}
\end{equation}
This comes from the fact that for $\epsilon \to 0$ we have: $W(i\epsilon, r+i \Delta) \sim \frac{\alpha_x}{-i\epsilon} W_y(r+i\Delta)$. Because we expect a real result, we ought to have numerically:
\begin{equation}
\begin{split}
    \opR_{x,y} \left\{ \gamma_t(x)\gamma_t(y) W(x,y)\right\}
    = \gamma_t(0)^2 \alpha_{xy}
    + \iint_{[\lambda_0, \lambda^*]^2}  \gamma_t(u) \gamma_t(v) \rho(u,v) \dd u \dd v\\
    +  \gamma_t(0) \int_{\lambda_0}^{\lambda^*} \dd r \gamma_t(r) 
    \lim_{\Delta \to 0^+} \lim_{\epsilon \to 0^+} \frac{\epsilon}{ \pi} \Re \biggl\{ W(i \epsilon, r - i \Delta) - W(i \epsilon, r + i \Delta) \biggr\}
\end{split}
\end{equation}
% Actually, as we expect $W_y(r-i \Delta) = \text{Conj}(W_y(r+i \Delta))$, we have:
%\begin{equation}
%    \epsilon \Re W(i \epsilon, r - i \Delta) \sim
%    \alpha_x \Re \left[ i W_y(r-i\Delta) \right]
%    \sim \alpha_x \Re \text{Conj} \left[ -i W_y(r+i\Delta) \right]
%    \sim - \epsilon \Re W(i \epsilon, r + i \Delta) 
%\end{equation}
%Hence we can simply write down:
%\begin{equation}
%\begin{split}
%    \opR_{x,y} \left\{ \gamma_t(x)\gamma_t(y) W(x,y)\right\}
%    = \gamma_t(0)^2 \alpha_{xy}
%    + \iint_{[\lambda_0, \lambda^*]^2}  \gamma_t(u) \gamma_t(v) \rho(u,v) \dd u \dd v\\
%    -  \frac{2 \gamma_t(0)}{\pi} \int_{\lambda_0}^{\lambda^*} \dd r \gamma_t(r) 
%    \lim_{(\epsilon, \Delta) \to 0^+}  \epsilon \Re W(i \epsilon, r + i \Delta)
%\end{split}
%\end{equation}

\paragraph{1-variable distributions in 2-variables functions} Finally, it can happen that the 2-variables functions $W(x,y)$ actually generates a distribution $\rho(u,v) = \rho_c(u,v) +  \mu(u)\delta(v-u)$ which may be the sum of a continuous measure $\rho_c(u,v)$ as described above, and another measure $\mu(u) \delta(v -u) = \delta(u-v) \mu(v)$.

\subsection{Additional heatmaps}
\begin{figure}[h!]
    \centering
        \includegraphics[width=1.0\textwidth]{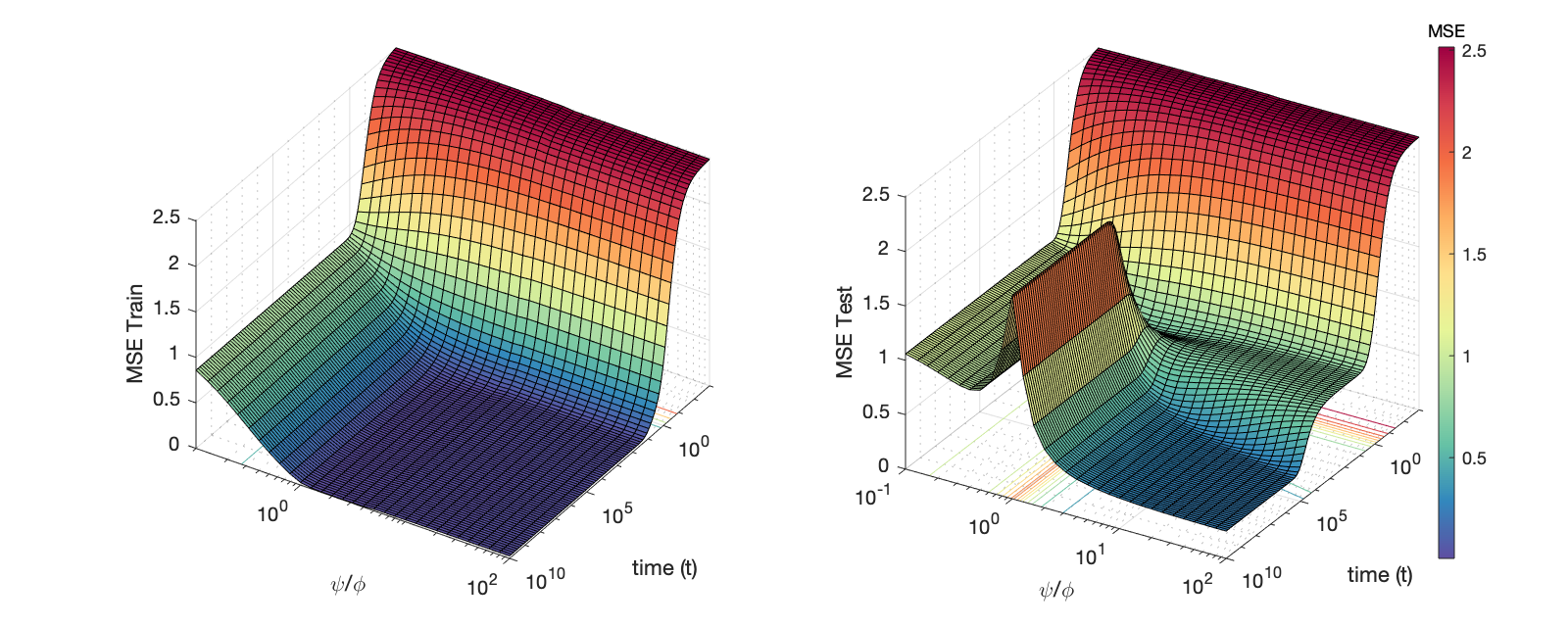}
    \caption{ 
        Analytical training error with parameters
        $(\mu,\nu,\phi,r,s,\lambda) = (0.5,0.3,3,2.,0.4,0.001)$
    }
    \label{fig:config2_3D}
\end{figure}

We provide additional heatmaps that complement those of Sect. \ref{results-and-insights}. Notice that all the heat-maps are always derived from a 3D mesh comprising $30\times 100$ points as in Fig. \ref{fig:config2_3D}.

\begin{figure}[h!]
    \centering
    \includegraphics[width=1.0\textwidth]{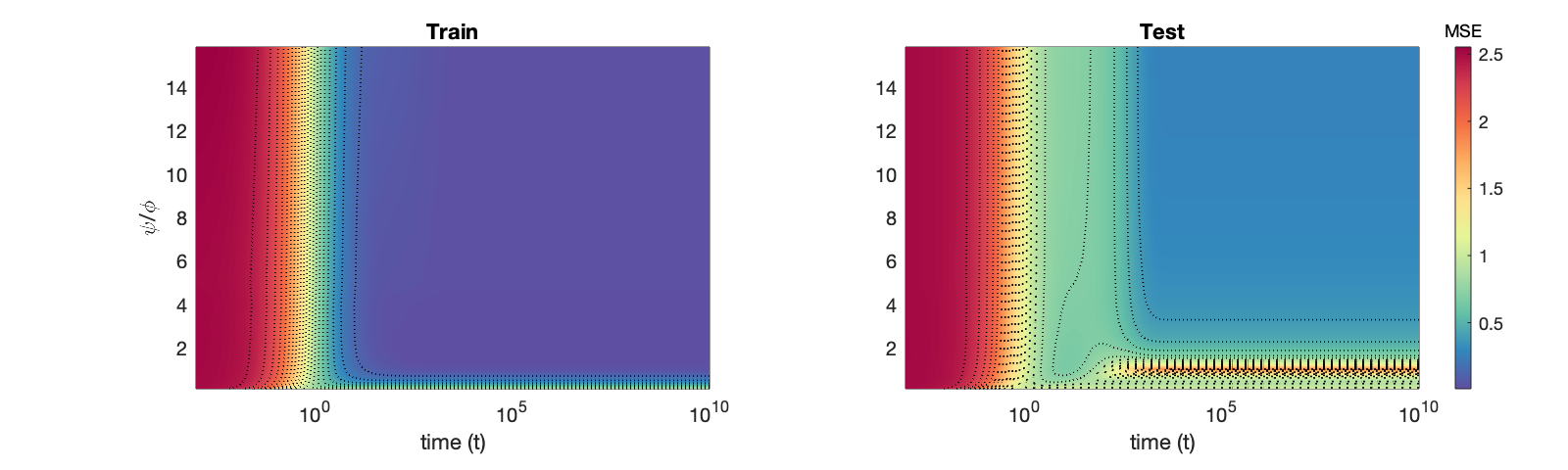}
    \caption{ 
    Analytical training error and test error evolution at fixed $\delta$ with parameters
    $(\mu,\nu,\phi,r,s,\delta) = (0.5,0.3,3,2.,0.4,0.001)$
    }
    \label{fig:time_double_descent_rescaled}
  \end{figure}

Instead of fixing $\lambda$, we can rescale it and fix $\delta = c \lambda$. As we have seen, the $\lambda$ parameter seems to affect the length  of the time scale on the first plateau. Rescaling it as seen in Fig. \ref{fig:time_double_descent_rescaled}, the interpolation threshold time scale becomes   constant in the over-parametrized regime at fixed $\delta$, and the results are consistent with what is observed empirically in \cite{nakkiran2019deep}. 

\begin{figure}[h!]
    \centering
    \includegraphics[width=1.0\textwidth]{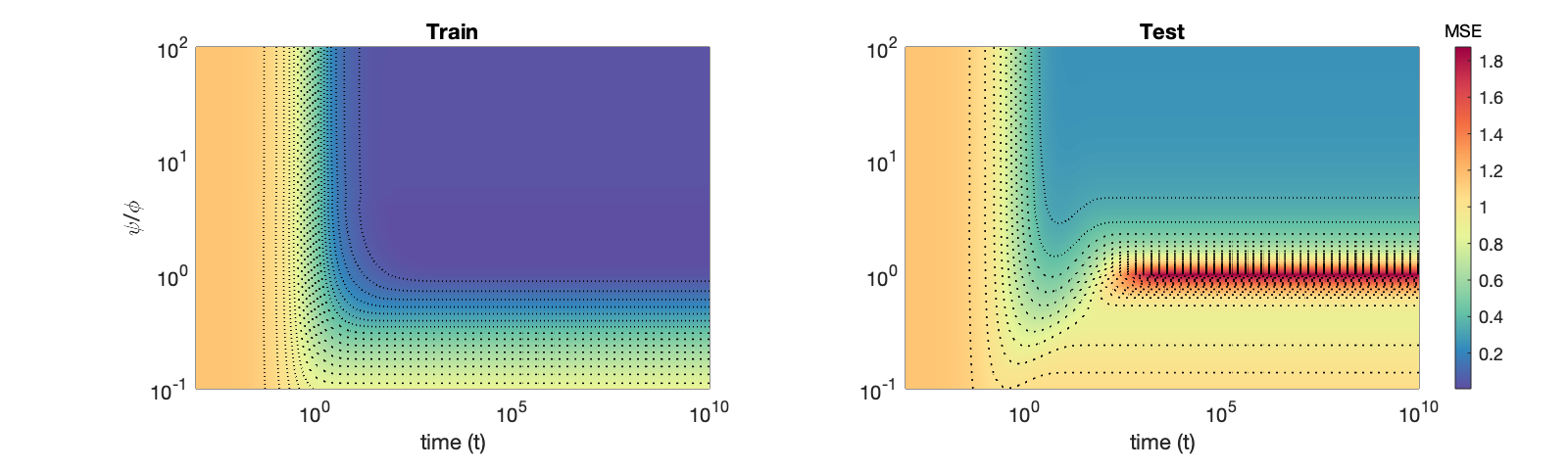}
    \caption{ 
    Analytical training error and test error evolution with parameters
    $(\mu,\nu,\phi,r,s,\lambda) = (0.5,0.3,3,0,0.4,0.001)$
    }
    \label{fig:config2_2D_r0}
  \end{figure}

We notice also that under the configuration in Fig. \ref{fig:config2_2D_r0} where $r=0$ (the noise of the second layer vanishes), the second plateau seems to vanish with the test error. 

\begin{figure}[h!]
    \centering
    \includegraphics[width=1.0\textwidth]{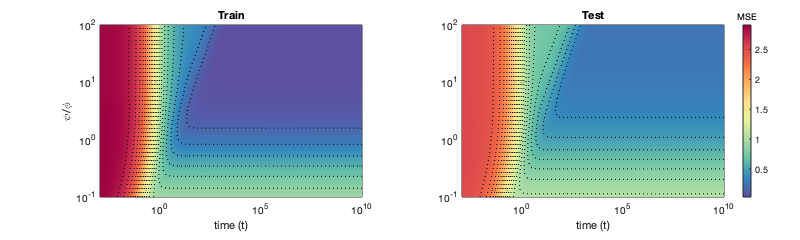}
    \caption{ 
    Analytical training error and test error evolution with parameters
    $(\mu,\nu,\phi,r,s,\lambda) = (0.5,0.3,3,0,0.4,0.1)$
    }
    \label{fig:config2_2D_big_lambda}
  \end{figure}

One of the effects of a large $\lambda$ is that it removes the double descent on the test error, which is consistent with the description in \cite{mei2020generalization}. Another effect is that it seems to add an additional "two-stage decrease" in the training error as can be seen in Fig. \ref{fig:config2_2D_big_lambda} and also in the experiments in Figs. \ref{fig:config2_train_100}, \ref{fig:epochwise-double-descent}. 

\begin{figure}[h!]
    \centering
    \includegraphics[width=1.0\textwidth]{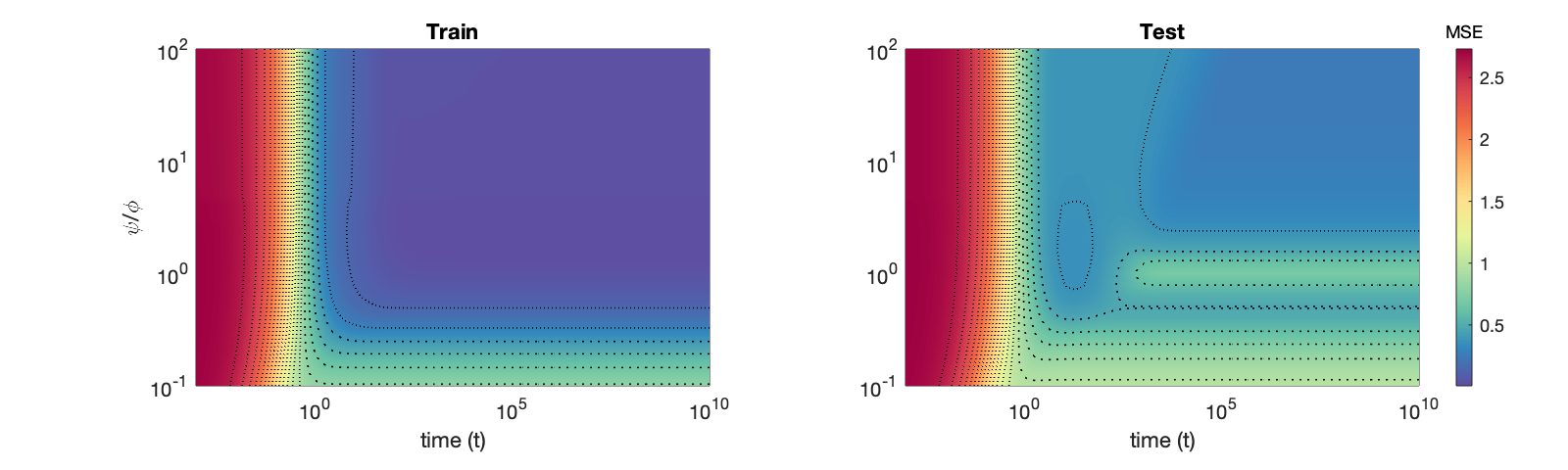}
    \caption{ 
    Analytical training error and test error evolution with parameters corresponding to $\sigma(x) = \tanh(x)$ with
    $(\mu,\nu,\phi,r,s,\lambda) = (0.61,0.15,3,0,0.4,0.001)$
    }
    \label{fig:config8_2D}
\end{figure}

\begin{figure}[h!]
    \centering
    \includegraphics[width=1.0\textwidth]{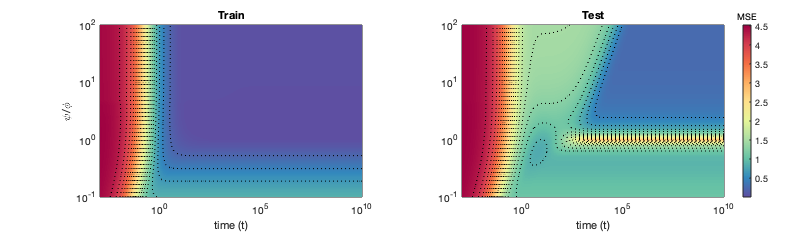}
    \caption{ 
        Analytical training error and test error evolution with parameters corresponding to $\sigma(x) = \tanh(5x)$ with
        $(\mu,\nu,\phi,r,s,\lambda) = (0.79,0.47,3,2,0.4,0.001)$
    }
    \label{fig:config9_2D}
\end{figure}

Note that the previous figures are perfomed for the activation function $\sigma(x) = \text{Relu}(x) - \frac{1}{\sqrt{2\pi}}$ while Figs. \ref{fig:config8_2D} and \ref{fig:config9_2D} are displayed other activation functions, $\sigma(x) = \tanh(x)$ and $\sigma(x) = \tanh(5x)$. We can see that the epoch-wise structures are more marked when the slope of the activation function is bigger in the second case.

  \begin{figure}[h!]
    \centering
    \includegraphics[width=1.0\textwidth]{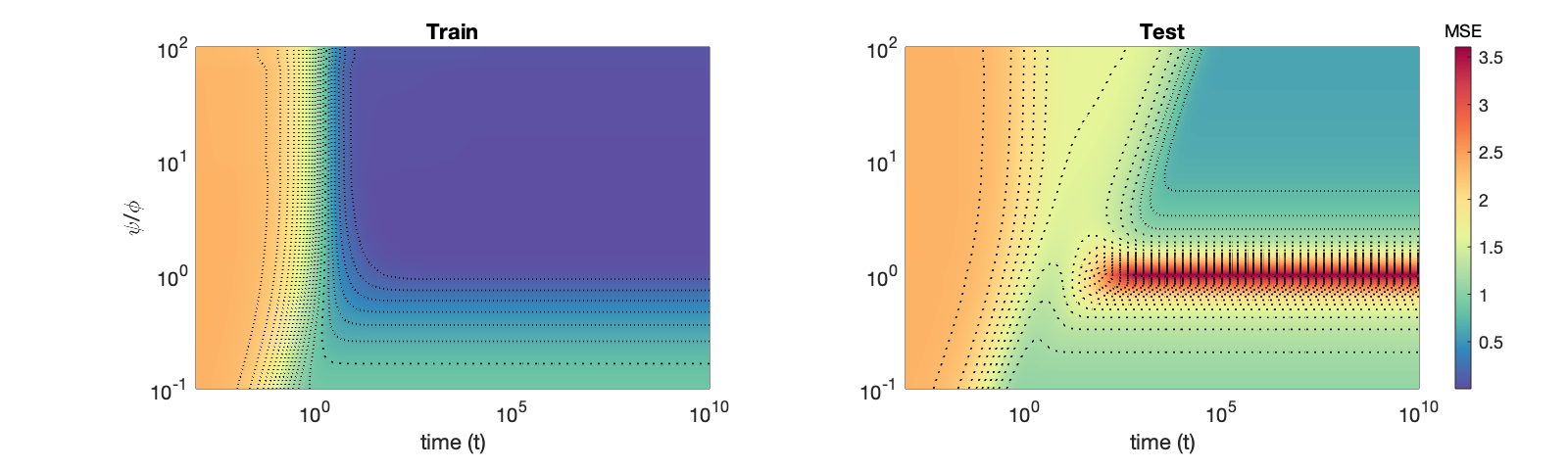}
    \caption{ 
    Analytical training error and test error evolution with parameters
    $(\mu,\nu,\phi,r,s,\lambda) = (0.5,0.3,0.5,2,0.1,0.003)$
    }
    \label{fig:config6_2D}
  \end{figure}


\subsection{Comparison with experimental simulations}

\begin{figure}[h!]
    \centering
    \subfloat{
        \includegraphics[width=0.45\textwidth]{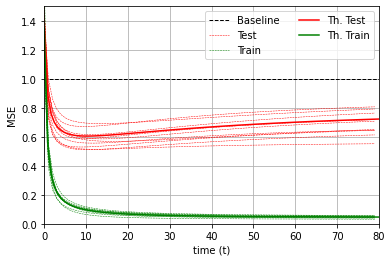}
    }
    \subfloat{
        \includegraphics[width=0.45\textwidth]{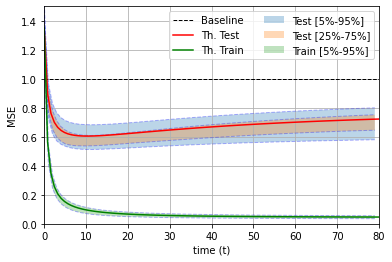}
    }
    \caption{ 
        Analytical training error and test error profile with parameters
        $(\mu,\nu,\phi,\psi,r,s,\lambda) = (0.5,0.3014,1.4,1.8,1.0,0,0.01)$
        compared to 10 experimental runs ($\sigma = \text{Relu} - \frac{1}{\sqrt{2\pi}}$) with $d=200$ and $\dd t=0.01$
    }
    \label{fig:config0_200}
\end{figure}

\begin{figure}[h!]
    \centering
    \subfloat{
        \includegraphics[width=0.45\textwidth]{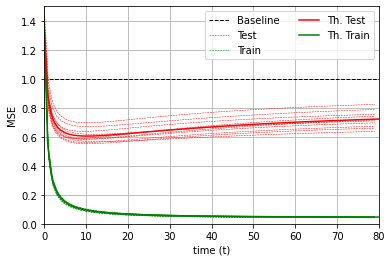}
    }
    \subfloat{
        \includegraphics[width=0.45\textwidth]{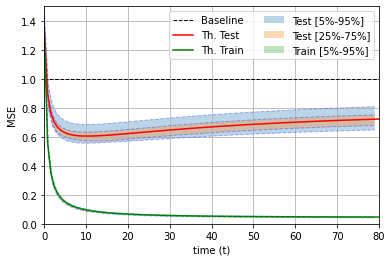}
    }
    \caption{ 
        Analytical training error and test error profile with parameters
        $(\mu,\nu,\phi,\psi,r,s,\lambda) = (0.5,0.3014,1.4,1.8,1.0,0,0.01)$
        compared to 10 experimental runs ($\sigma = \text{Relu} - \frac{1}{\sqrt{2\pi}}$) with $d=1000$ and $\dd t=0.01$
    }
    \label{fig:config0_1000}
\end{figure}

We have already shown on figure \ref{fig:tripledescent} in Sect. \ref{results-and-insights}  that the analytical formulas for the training and generalization errors match the experimental curves in the limit of $t\to +\infty$. Here we provide additional evidence that this is also the case for the whole time-evolution in Figs. \ref{fig:config0_200} and \ref{fig:config0_1000} as the dimension $d$ increases.

\begin{figure}[h!]
    \centering
        \includegraphics[width=0.6\textwidth]{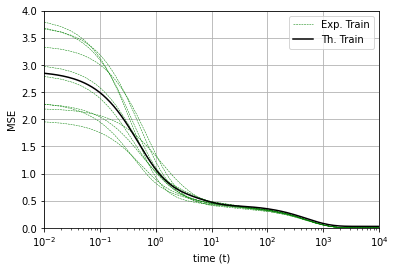}
    \caption{ 
        Analytical training error with parameters
        $(\mu,\nu,\phi,\psi,r,s,\lambda) = (0.5,0.3,300,3,2,0.4,0.1)$
        compared to 10 experimental runs ($\sigma = \text{Relu} - \frac{1}{\sqrt{2\pi}}$) with $d=100$ and $\dd t=0.01$ 
    }
    \label{fig:config2_train_100}
\end{figure}

\begin{figure}[h!]
    \centering
        \includegraphics[width=0.6\textwidth]{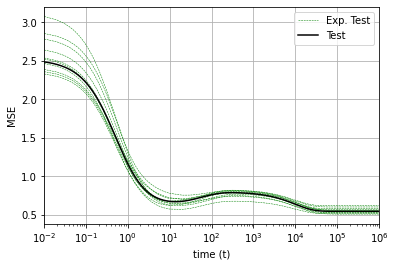}
    \caption{ 
        Analytical test error with parameters
        $(\mu,\nu,\phi,\psi,r,s,\lambda) = (0.5,0.3,6,3,2,0.4,0.0001)$
        compared to 10 experimental runs with $d=100$ and $\dd t=0.01$ for $0 \leq t \leq 10^4$ and $\dd t=0.1$ for $10^4 \leq t \leq 10^6$
    }
    \label{fig:epochwise-double-descent}
\end{figure}

In \ref{fig:config2_train_100}, \ref{fig:epochwise-double-descent}, we can see that the epoch-wise descent structures of the training error and test error can be captured correctly experimentally for long time. Note that we have taken $d=100$ small enough to be able to run these experiments for such a long timescale.

